\documentclass[]{bytedance_seed}

\usepackage[toc,page,header]{appendix}

\usepackage{minitoc}
\usepackage{booktabs}
\usepackage{makecell}

\usepackage{amsmath,amsfonts,bm}

\def\eqref#1{equation~\ref{#1}}

\def\1{\bm{1}}

\DeclareMathAlphabet{\mathsfit}{\encodingdefault}{\sfdefault}{m}{sl}
\SetMathAlphabet{\mathsfit}{bold}{\encodingdefault}{\sfdefault}{bx}{n}

\usepackage{colortbl}
\usepackage{xcolor}
\definecolor{mygray}{gray}{.9}
\definecolor{goldenrod}{RGB}{245,245,220}
\newlength\savewidth
\newcolumntype{a}{>{\columncolor{mygray}}c}
\usepackage{fontawesome5}
\usepackage{booktabs}
\usepackage{makecell}
\usepackage{fontawesome5}
\usepackage{lipsum}
\usepackage{comment}
\usepackage{multirow}
\usepackage{wrapfig}
\usepackage{algorithm}
\usepackage{algorithmic}
\usepackage{xfrac}  

\usepackage{graphicx}
\usepackage{color}

\usepackage{amsthm}
\definecolor{darkgreen}{rgb}{0,0.7,0}

\definecolor{mygraytext}{gray}{.75}

\title{UniEdit-I: Training-free Image Editing for Unified VLM via Iterative Understanding, Editing and Verifying}

\author[1,2,\bullet]{Chengyu Bai}
\author[1]{Jintao Chen}
\author[1]{Xiang Bai}
\author[1]{Yilong Chen}
\author[2,\star]{\\Qi She}
\author[1,\star]{Ming Lu}
\author[1, \dagger]{Shanghang Zhang}

\affiliation[1]{Peking University}
\affiliation[2]{ByteDance}

\contribution[\bullet]{Work done during ByteDance Internship, supervised by Qi She}
\contribution[\star]{Project leaders}
\contribution[\dagger]{Corresponding author}

\definecolor{highlight}{RGB}{220,230,241}
\definecolor{badcase}{RGB}{255,200,200}

\abstract{
While Unified Vision-Language Models promise to synergistically combine the high-level semantic understanding of vision-language models with the generative fidelity of diffusion models, current editing methodologies remain fundamentally decoupled and open-loop—performing static, pre-defined transformations without dynamic feedback between semantic interpretation and visual generation. A central limitation stems from the representation gap: understanding typically leverages high-level, language-aligned encoders, whereas generation relies on low-level, pixel-space autoencoders, resulting in misaligned feature spaces. To bridge this gap, Recent advances such as Representation Autoencoders and BLIP3-o advocate performing diffusion-based modeling directly in high-level features from pretrained semantic encoders. We find editing in the semantic latent space modifies conceptual representations rather than pixels, ensuring intermediates that are both semantically coherent and visually plausible. Building on this insight, We propose \textbf{UniEdit-I}, the first training-free, closed-loop image editing framework that operates entirely within the semantic latent space of a unified VLM by introducing an \textbf{Understanding–Editing–Verifying (UEV)} loop:
(1) Understanding: parses the source image and editing instruction into a structured source prompt and a minimal target specification;
(2) Editing: applies dynamic semantic offsets, with a configurable feedback weighting mechanism that adaptively modulates editing intensity based on real-time alignment feedback;
(3) Verifying: leverages the VLM’s own multimodal reasoning capability to evaluate the intermediate output along multiple semantic dimensions and trigger early stopping or refinement. By transforming the VLM from a post-hoc evaluator into an in-process conductor, UniEdit-I establishes the first semantics-driven, self-correcting closed-loop image editing pipeline. Evaluated on GEdit-Bench, UniEdit-I achieves state-of-the-art performance without any fine-tuning or architectural modifications, and even surpasses several large-scale pre-trained editors. 
}

\date{\today}
\correspondence{Shanghang Zhang at \email{shanghang@pku.edu.cn}}

\begin{document}
\maketitle

\begin{figure}
    \centering  
    \captionsetup{type=figure}
    \includegraphics[width=\textwidth]{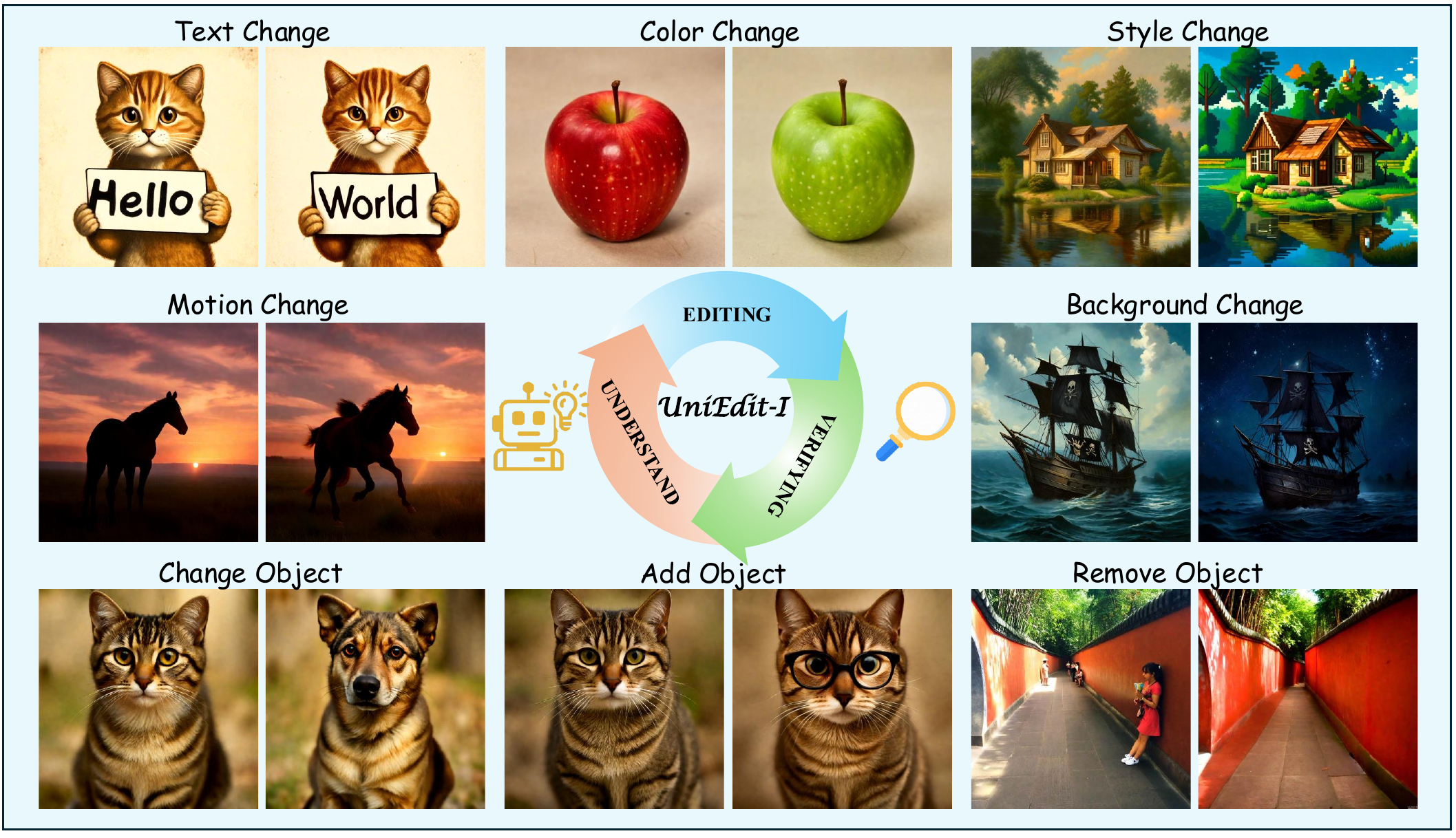}
    \captionof{figure}{The illustration of UniEdit-I. We introduce a novel training-free framework named UniEdit-I to enable the unified VLM with image editing capability via three iterative steps: \textbf{understanding, editing, and verifying}.}
    \label{fig:teaser}  
\end{figure}

\section{Introduction}
\label{sec:intro}
The core challenge in image editing lies in reliably translating high-level semantic instructions into visually faithful and coherent modifications. Although the proposal of unified vision-language models (Unified VLMs) aims to integrate VLMs and diffusion models, leveraging the former for high-level semantic understanding and the latter for precise generation, thereby enabling synergistic mutual enhancement, current editing methods still operate in a functionally decoupled, open-loop framework. These approaches perform static, pre-defined transformations along fixed trajectories, lacking dynamic, feedback-driven integration between semantic interpretation and visual generation.

\begin{figure}[ht]
    \centering
    \includegraphics[width=\textwidth]{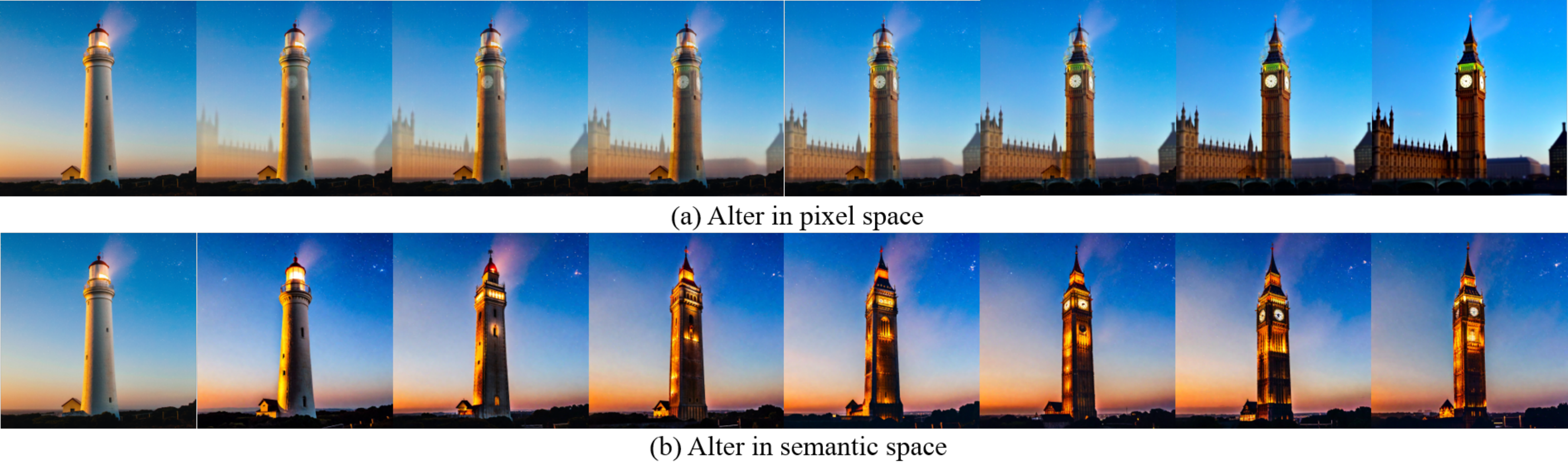}
    \caption{(a) In pixel space, intermediate outputs exhibit a superposition of source and target content, resulting in visible ghosting and unnatural transitions.(b) In semantic space, intermediate states are clean and realistic, with coherent structure and no artifacts, leading to a natural and faithful final result.}
    \label{fig:semantic}
\end{figure}

In addition, A key challenge in Unified VLMs lies in the representation gap: understanding typically benefits from high-level, language-aligned semantic encoders (CLIP~\cite{clip}, SigLIP~\cite{siglip}), while generation traditionally relies on low-level, pixel-preserving autoencoders like VAEs~\cite{vae}. This misalignment between the feature spaces of understanding and generation further exacerbates the decoupling of semantic interpretation and visual generation, as the two processes operate on disjoint representations that lack inherent semantic consistency. Recent advances such as Representation Autoencoders~\cite{rae} (RAEs) and BLIP3-o~\cite{blip3o} advocate performing diffusion-based modeling directly in high-level features from pretrained semantic encoders, thereby unifying visual understanding and generation in a shared semantic latent space. Notably, we find that editing in this semantic latent space modifies the underlying conceptual representation rather than directly manipulating pixel values, ensuring that every intermediate state remains both semantically coherent and visually plausible (see Figure~\ref{fig:semantic}).

Building on this, We propose UniEdit-I, the first training-free, closed-loop image editing framework, that enable image editing directly in the semantic latent space via three iterative steps: \textbf{understanding}, \textbf{editing}, and \textbf{verifying}. 

\textbf{Understanding.} UniEdit-I first analyze the source image to generate a structured source prompt \(C_{\text{src}}\) and a scene graph \(\mathcal{G}\). The editing instruction is then parsed to produce a minimal modification, yielding a target prompt \(C_{\text{tar}}\).

\textbf{Editing.} UniEdit-I applies a configurable feedback weighting mechanism to dynamically modulate editing intensity, which operates entirely in the semantic latent space. 

\textbf{Verifying.} UniEdit-I evaluate the intermediate image by the VLM to extract multi-dimensional semantic feedback.

UniEdit-I’s fundamental insight is simple yet profound: if a model can understand an instruction and generate an image, it should also be able to judge whether the process of generation faithfully follows that intent—and use that judgment to steer the process itself. We do not treat this ability as a final evaluation; we embed it as the core driver of editing. In doing so, we unlock a new class of image editors that are more robust, more controllable, and more aligned with human intention.

Evaluated on GEdit-Bench, UniEdit-I achieves state-of-the-art performance without any fine-tuning or architectural modifications, surpassing even large-scale pre-trained editors. Ablations confirm that both the shift to semantic space and the closed-loop feedback are critical to performance; the configurable weighting further enhances adaptability across diverse user intents.

Our contributions are:
\begin{itemize}
\item We propose the first training-free, closed-loop image editing framework, which leverages real-time VLM feedback to dynamically guide the editing trajectory without requiring architectural modifications or fine-tuning;
\item We design a configurable feedback weighting mechanism that enables task-adaptive control over editing intensity, thereby enhancing robustness and user agency;
\item We show that high-quality semantic latent spaces (e.g., the CLIP space from unified VLMs) are not merely preferable for generation but also necessary for reliable closed-loop editing, establishing a new paradigm of "reflective" generative systems.
\end{itemize}

\section{Related Work}
\label{sec:relatedwork}

\subsection{Unified Vision-Language Models}
Recent years have witnessed rapid progress toward unified vision-language models (VLMs)—architectures that perform both visual understanding and generation within a single framework. Numerous unified VLMs have explored diverse architectural designs~\cite{emu2,emu3,metaqueries,bagel,mmada,huang2025illume+,omnigen2,lavit,dreamllm,vlgpt,puma,blip3o,lwm,chameleonteam,anole,stnergen,ugen}. Early approaches relied exclusively on autoregressive (AR) modeling over discrete visual tokens—such as those from VQGAN~\cite{vqgan}, VQ-IMG~\cite{vqimg}, and SBER-MoVQGAN~\cite{sber}. However, suffered from poor fidelity and limited detail due to information bottlenecks. In response, recent unified VLMs now adopt hybrid architectures. These retain an AR component for high-level semantic planning but delegate pixel-level synthesis to a dedicated generative head. This design allows the model to preserve the strong multimodal reasoning capabilities of understanding backbone while achieving high-quality, diverse image generation, and has become the new standard.

Despite this architectural convergence, a fundamental challenge remains: the visual representations optimal for understanding and generation are inherently divergent. Understanding tasks benefit from semantic, language-aligned encoders such as SigLIP~\cite{siglip}, EVA-CLIP~\cite{evaclip}, and UNIT~\cite{unit}. In contrast, traditional generative models rely on pixel-preserving autoencoders like VAEs~\cite{vae}. Representation Autoencoders (RAEs)~\cite{rae} offer an elegant resolution to this tension. By pairing a frozen, pretrained semantic encoder with a lightweight trainable ViT decoder, RAE enables diffusion transformers to operate directly in this semantic latent space. 

\subsection{Image Editing with Diffusion Models}
Driven by breakthroughs in diffusion-based text-to-image models\cite{rombach2022high,esser2024scaling,xie2024sana}, many methods now focus on diffusion-based image editing. Most rely on inversion techniques\cite{song2020denoising,meng2021sdedit,rout2024semantic} to map an image to noise before re-sampling under new conditions. To bypass inversion, alternatives such as attention manipulation\cite{hertz2022prompt,tumanyan2023plug} and optimization-based editing\cite{kim2022diffusionclip,kwon2022diffusion} have been explored. Recently, FlowEdit\cite{kulikov2024flowedit} proposed an inversion-free trajectory in pixel or VAE space, enabling visible intermediate edits. However, it operates in an open-loop manner: edits are applied without real-time semantic feedback; Artifacts like ghosting or deformation in these intermediate states undermine visual consistency and hinder reliable closed-loop control by VLMs.

\subsection{Unified VLM‑based Editing}
While unified VLMs such as GPT-4o~\cite{openai}, Step1X-Edit~\cite{gedit}, and BAGEL~\cite{bagel} achieve strong in context editing performance, they typically rely on large scale training on paired edit data. In contrast, we propose the first training free closed loop editing framework for unified VLMs. By constructing editing trajectories in CLIP semantic space and introducing an Understanding Editing Verifying loop, our method uses frozen VLMs as active agents of the editing process, enabling adaptive self correcting image manipulation without any parameter updates.

\section{Preliminaries}
\subsection{The FlowEdit Algorithm} \label{subsec:FlowEdit_Algorithm}
FlowEdit~\cite{kulikov2024flowedit} introduces an inversion-free approach to image editing by directly constructing a continuous trajectory from the source to the target distribution in pixel or VAE latent space. FlowEdit starts the ODE integration from the source image latent $Z_0^{\text{src}}$ at timestep $t_{n_{\max}}$, setting the initial state as:$Z_{t_{n_{\max}}}^{FE} = Z_0^{\text{src}}$.

At each denoising step $t_i$, FlowEdit constructs a pair of noisy latents—$Z_{t_i}^{\text{src}}$ (from $Z_0^{\text{src}}$ and noise) and $Z_{t_i}^{\text{tar}}$ (aligned with it)—and computes the velocity difference:
\begin{gather}
    Z_{t_i}^{tar} = Z_{t_i}^{FE} + Z_{t_i}^{src} - Z_0^{src},\\
    \Delta V(t_i) = V(Z_{t_i}^{\text{tar}}, t_i, C_{\text{tar}}) - V(Z_{t_i}^{\text{src}}, t_i, C_{\text{src}}).
\end{gather}
Then updates the editing path via Euler integration:
\[
Z_{t_{i-1}}^{FE} = Z_{t_i}^{FE} + (t_{i-1} - t_i) \Delta V(t_i).
\]
The final state of the path at $t_0 = 0$ is the fully edited, noise-free image: $Z_{t_0}^{FE}$. More importantly, all intermediate states along its constructed editing path, $Z_t^{FE}$, are visible images. This observability opens the door to closed-loop control. Yet observability does not guarantee semantic reliability. Intermediate outputs in pixel or VAE space often suffer from artifacts—ghosting, object deformation, making it difficult for VLMs to generate stable, accurate feedback. As a result, even with visible trajectories, closed-loop editing remains ineffective.

\subsection{BLIP3-o Architecture} \label{subsec:BLIP3o_Architecture}
The text-to-image generation process of BLIP3-o consists of three stages. It first translates a text prompt into a set of conditional visual features, then generates a corresponding CLIP image feature using a diffusion model, and finally decodes this feature into a high-resolution image.

\textbf{Conditional Feature Generation:} Given a text prompt $P$, vision-language model (VLM) first produces text embeddings $C = E_{text}(P)$. These are concatenated with a series of learnable query embeddings, $Q_{query}$, and processed by the VLM's backbone $\Phi_{VLM}$, to produce conditional visual features, $Q_{cond}$. This can be expressed as:
$$Q_{cond} = \Phi_{VLM}([C ; Q_{query}])[-L_Q:, :]$$
where $L_Q$ is the length of the query embeddings $Q_{query}$.

\textbf{CLIP Feature Generation:} A Diffusion Transformer (DiT), $D_{\theta}$, then generates a CLIP image feature, $\hat{X}_1$, conditioned on $Q_{\text{cond}}$. This process starts from a random noise vector, $X_0 \sim \mathcal{N}(0, I)$, and solves the flow-matching ODE $\frac{dX_t}{dt} = D_{\theta}(X_t, Q_{\text{cond}}, t)$ using $T$ discrete steps via numerical integration. Denoting the timesteps as $t_0 < t_1 < \dots < t_T$ with $t_0=0$ and $t_T=1$, the iterative update is:
\begin{equation}
X_{t_{i-1}} = X_{t_i} + (t_{i-1} - t_i) \cdot D_{\theta}(X_{t_i}, Q_{\text{cond}}, t_i),
\end{equation}
where $i$ runs from $T$ down to $1$. The final CLIP feature is obtained after $T$ steps:
$\hat{X}_1 = X_{t_0}$.

\textbf{Image Decoding:} Finally, a separate, pre-trained image decoder, $\mathcal{G}_{dec}$, synthesizes the final pixel image, $I_{pixel}$, from the generated CLIP feature, $\hat{X}_1$:
$I_{pixel} = \mathcal{G}_{dec}(\hat{X}_1)$.

\section{Method}
\label{sec:method}

\begin{figure*}[t]
  \centering
  \includegraphics[width=\textwidth]{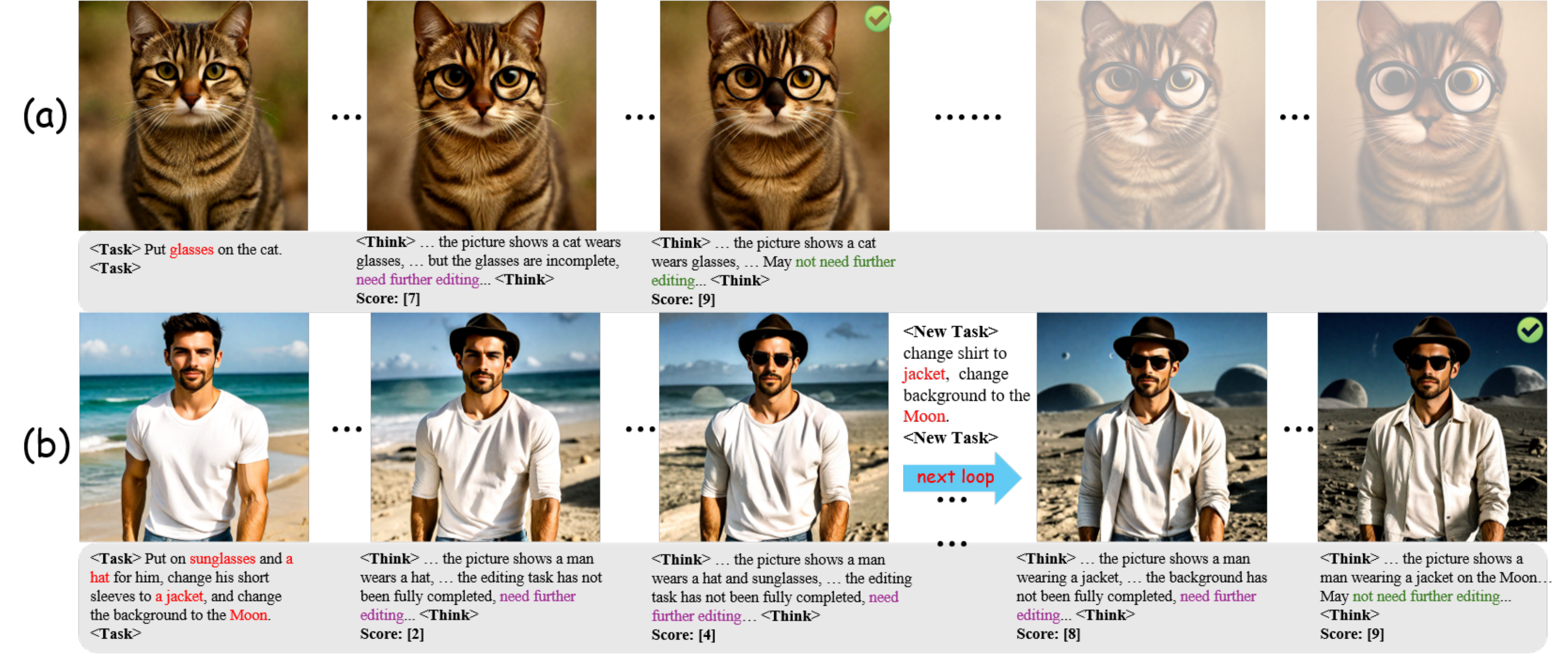}
  \caption{Edited Image Verification. Evaluates alignment between the target prompt and the intermediate edited image, provides automatic consistency scores and corrective feedback, and determines whether to stop early or continue the editing loop.}
  \label{fig:verify}
\end{figure*}

We present \textbf{UniEdit-I}, the first training-free, closed-loop image editing framework that leverages the semantic latent space of a unified vision-language model (VLM) to enable real-time, self-correcting edits. Unlike prior methods that operate in pixel or VAE space, UniEdit-I constructs and steers the editing trajectory entirely within the \textit{CLIP feature space} of a unified VLM (e.g., BLIP3-o~\cite{blip3o}). Our approach is governed by an iterative \textbf{Understanding–Editing–Verifying (UEV)} loop, illustrated in Figure~\ref{fig:verify}, which dynamically adapts editing intensity based on continuous semantic feedback. Below, we detail each stage.

\subsection{Understanding: Structured Prompt Generation}
\label{subsec:understanding}

\begin{figure*}[t]
  \centering
  \includegraphics[width=0.9\textwidth]{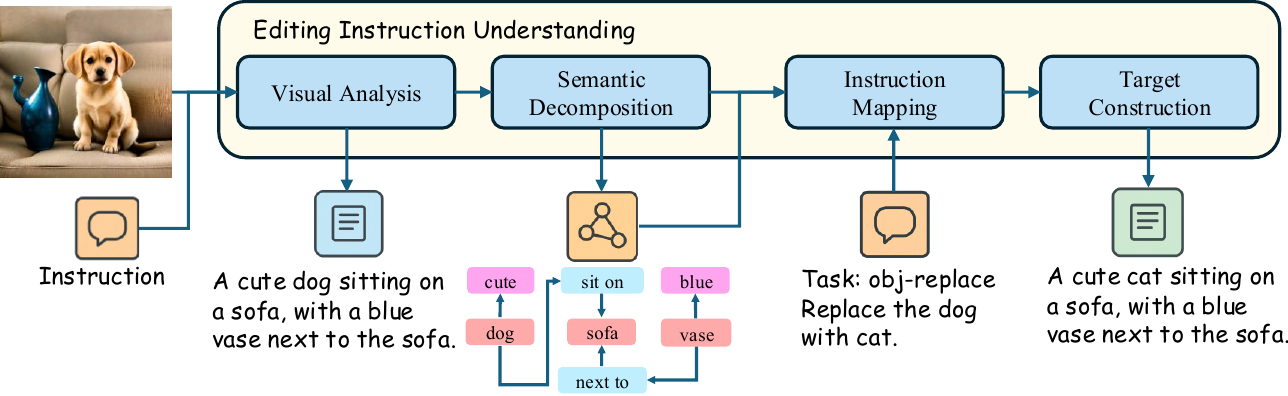}
  \caption{Structured Prompt Generation pipeline. Visual analysis→semantic decomposition→instruction mapping→target construction.}
  \label{fig:understand}
\end{figure*}

In practice, users often provide only a target instruction without an explicit source description. To handle this, we first use the VLM to automatically generate a \textbf{source caption} $C_{\text{src}}$—a comprehensive, scene-aware textual description of $I_{\text{src}}$, conditioned on the edit type $\tau$ (e.g., ``attribute change'', ``object replacement'') to emphasize potentially editable elements. As shown in Figure\ref{fig:understand}, given $I_{\text{src}}$ and $q$, we prompt the language backbone with a system instruction tailored to $\tau$ and obtain two structured outputs in a single forward pass:
\begin{itemize}
\item \textbf{Source caption} $C_{\text{src}}$: A comprehensive, scene-aware description of $I_{\text{src}}$, generated under guidance from $\tau$ to include all potentially editable elements.
\item \textbf{Scene graph} $\mathcal{G} = (\mathcal{V}, \mathcal{E})$: A structured representation of objects $\mathcal{V}$ (with attributes) and their spatial relationships $\mathcal{E}$ (e.g., ``cat on sofa'').
\end{itemize}
We then parse $q$ to identify minimal token-level modifications to $C_{\text{src}}$, yielding the target caption $C_{\text{tar}}$. Simultaneously, we update $\mathcal{G}$ to $\mathcal{G}_{\text{tar}}$ by modifying relevant nodes or edges, while preserving all unmodified relationships. The triplet $\{C_{\text{src}}, C_{\text{tar}}, \mathcal{G}_{\text{tar}}\}$ serves as structured semantic supervision for both editing and verification.

\subsection{Semantic Trajectory Construction in semantic Space}
\label{subsec:trajectory}

We reinterpret the FlowEdit~\cite{kulikov2024flowedit} framework within the semantic latent space of BLIP3-o, effectively replacing its pixel- or VAE-based trajectory with one that operates directly on CLIP image features. This design is made possible by the architecture of BLIP3-o (Sec.~\ref{subsec:BLIP3o_Architecture}), whose generative process explicitly produces and manipulates CLIP features as an intermediate representation before final decoding. Specifically, BLIP3-o first conditions on text to generate visual queries $Q_{\text{cond}}$, then uses a diffusion transformer $D_\theta$ to synthesize a CLIP feature $\hat{X}_1$, which is subsequently decoded to pixels via $\mathcal{G}_{\text{dec}}$. This decoupling allows us to treat the CLIP feature as a \textit{native, editable canvas}.

\begin{figure}[ht]
    \centering
    \includegraphics[width=0.6\textwidth]{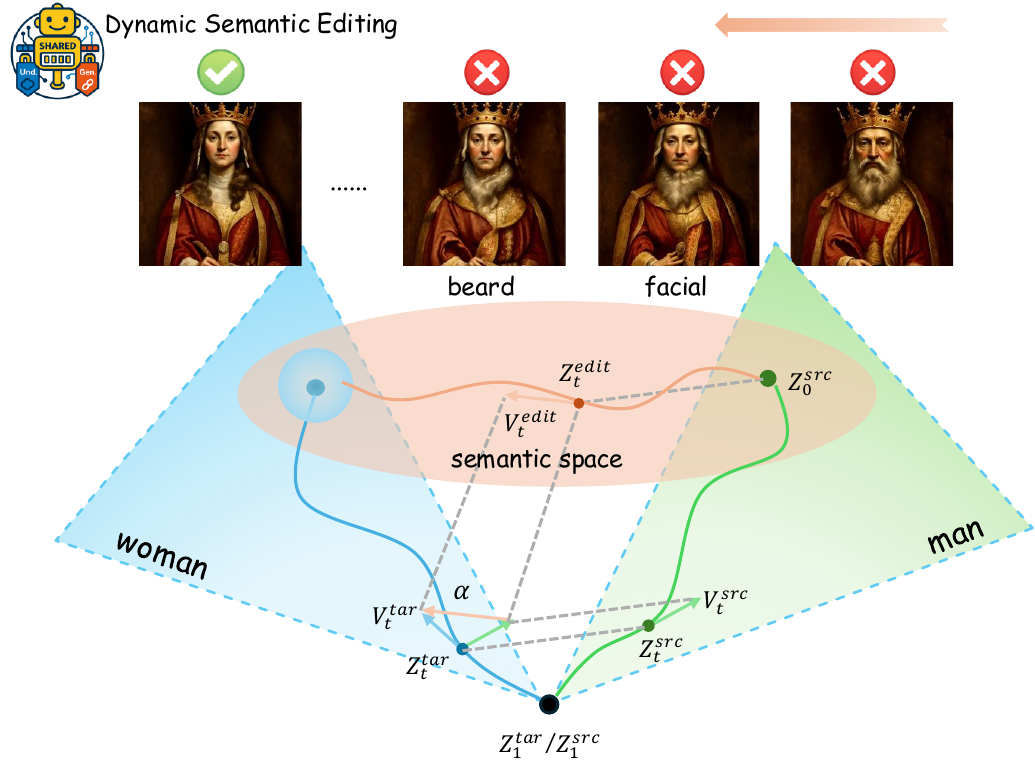}
    \caption{ \textit{Semantic trajectory editing in CLIP space(right to left).} We build on FlowEdit’s $\Delta V_t$, but apply it adaptively using VLM feedback to avoid artifacts ($\times$) and stop at the correct target ($\checkmark$).}
    \label{fig:explain}
\end{figure}

Given a source image $I_{\text{src}}$, we extract its CLIP feature $Z_{\text{src}}$ using BLIP3-o’s frozen vision encoder. Our goal is to generate a modified CLIP feature $Z_{\text{tar}}$ that aligns with the target caption $C_{\text{tar}}$, while preserving irrelevant source attributes. To achieve this, we adopt FlowEdit’s inversion-free ODE formulation (Sec.~\ref{subsec:FlowEdit_Algorithm}), but replace its Iterative Update step(Sec.~\ref{subsec:BLIP3o_Architecture}) with an equivalent process in CLIP space. Starting from the initial state $Z_{t_{n_{\max}}}^{\text{UE}} = Z_{\text{src}} $,
we iterate backward in time. At each timestep $t_i$, we construct noise-shared probes analogous to FlowEdit:
\begin{align}
    Z_{\text{src}}(t_i) &= (1 - \lambda(t_i)) Z_{\text{src}} + \lambda(t_i) \epsilon(t_i), \\
    Z_{\text{tar}}(t_i) &= Z_{\text{edit}}(t_i) + Z_{\text{src}}(t_i) - Z_{\text{src}},
\end{align}
where $\epsilon(t_i) \sim \mathcal{N}(0, I)$ and $\lambda(t_i)$ is a noise schedule.
We then query the same diffusion transformer $D_\theta$ used in BLIP3-o’s CLIP feature generation stage to compute the semantic velocity difference
\begin{equation}
    \Delta{V}(t_i) = V(Z_{t_i}^{tar}, t_i, C_{tar}) - V(Z_{t_i}^{src}, t_i, C_{src}).
\end{equation}
The editing trajectory is updated via Euler integration:
\begin{equation}
    Z_{t_{i-1}}^{\text{UE}} = Z_{t_i}^{\text{UE}} + (t_{i-1} - t_i) \cdot \alpha_{t_i} \cdot \Delta V(t_i),
    \label{eq:adaptive_gain}
\end{equation}
where $\alpha_{t_i}$ is the adaptive gain introduced in Sec.~\ref{subsec:editing}. The final output is $Z_{\text{edit}} = Z_{t_0}^{\text{UE}}$, decoded by the $\mathcal{G}_{\text{dec}}$ into $I_{\text{out}}$.

\textbf{Key insight}: Because BLIP3-o’s CLIP features are both semantically rich and natively decodable, every intermediate $Z_t^{\text{UE}}$ yields a clean, artifact-free image—unlike pixel-space intermediates that suffer from ghosting or distortion (Fig.~\ref{fig:semantic}). This property is essential: it transforms the open-loop editing process into a \textit{verifiable, closed-loop trajectory}, enabling real-time VLM feedback at every step.

\subsection{Editing: Adaptive Semantic Flow with Dynamic Gain}
\label{subsec:editing}

FlowEdit~\cite{kulikov2024flowedit} applies a uniform offset $\Delta V(t_i)$ across a pre-defined editing window $[n_{\max}, n_{\min}]$, resulting in static editing intensity. While simple, this open-loop strategy lacks any mechanism to assess the current state of semantic alignment or visual fidelity—leading to uncontrolled editing that may over-modify when alignment is already achieved, or under-modify when further refinement is still needed.

\begin{figure}[ht]
    \centering
    \includegraphics[width=0.6\textwidth]{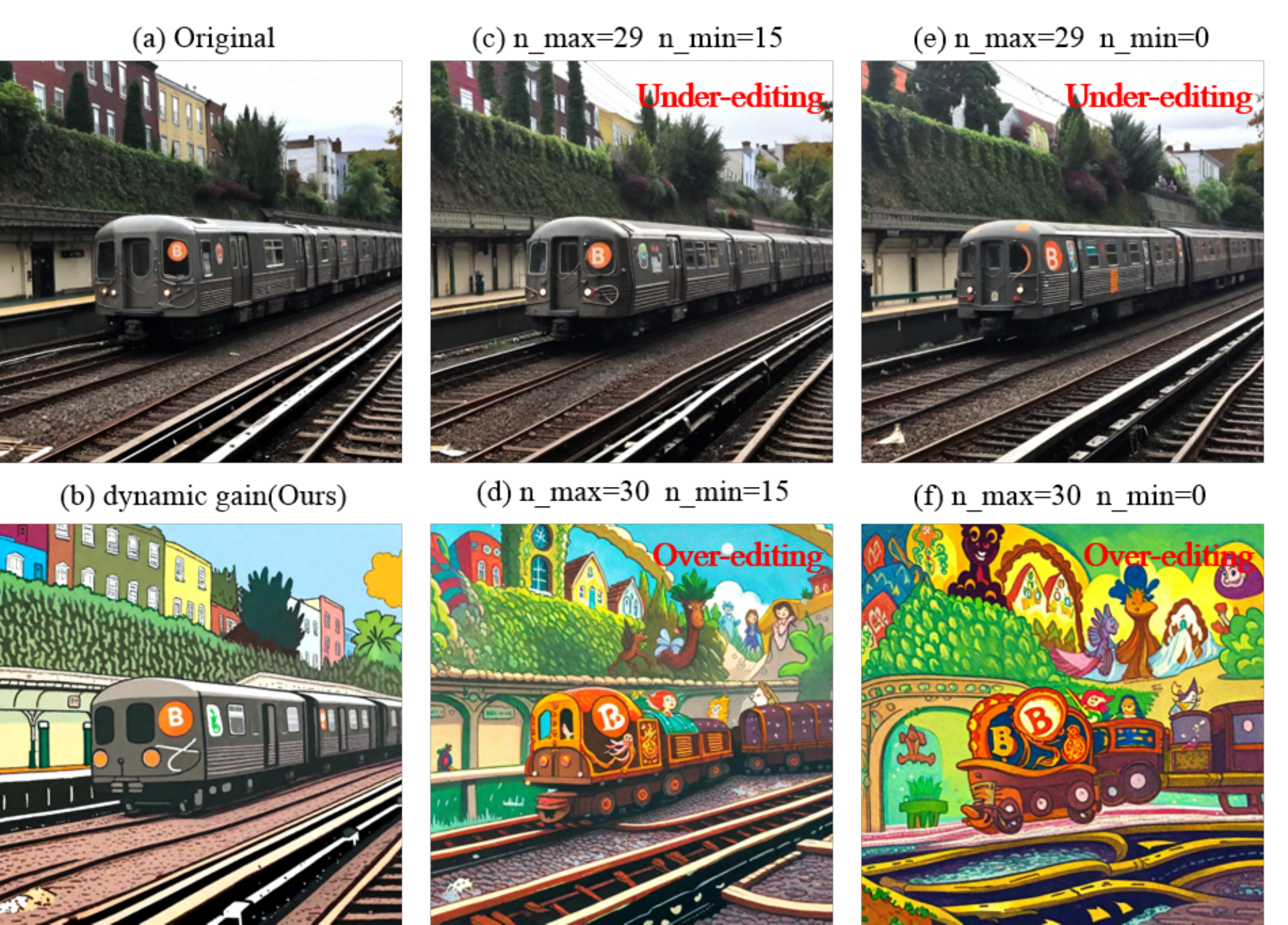}
    \caption{\textit{UniEdit-I outperforms FlowEdit variants by adapting both intensity and duration.} (a) Source image; (c–f) FlowEdit with fixed gain ($\alpha=1.0$) under different $[n_{\max}, n_{\min}]$ settings, all suffering from over- or under-editing; (b) Our method, using real-time feedback to stop early and preserve semantics. Only Ours achieves faithful, artifact-free editing without manual tuning.}
    \label{fig:adaptive gain}
\end{figure}

To address this, UniEdit-I introduces a \textbf{dynamic gain mechanism} that modulates editing strength based on real-time semantic progress. Specifically, every $k=5$ diffusion steps, we receive alignment feedback from the Verifying module (Sec.~\ref{subsec:verifying}) and compute an adaptive gain $\alpha_t$ for the next segment of editing. The gain is defined as:
\begin{equation}
    \alpha_t = \alpha_{\text{base}} \cdot \sigma(\kappa_1 \Delta s_t) \cdot (1 - p_t),
    \label{eq:alpha_t}
\end{equation}
where:
\begin{itemize}
    \item $\alpha_{\text{base}} = 1.0$ sets a nominal editing strength;
    \item $\Delta s_t = s_t - s_{\text{prev}}$ is the improvement in global semantic alignment since the last feedback point;
    \item $p_t \in [0,1]$ is a task-completion score reflecting how closely the current output satisfies the instruction;
    \item $\sigma(\cdot)$ is a sigmoid function ($\kappa_1 = 15$) that amplifies gain when alignment is improving ($\Delta s_t > 0$) and suppresses it otherwise.
\end{itemize}
This gain is applied uniformly to all integration steps within the upcoming segment $[t, t-k)$ in Equation \ref{eq:adaptive_gain}.

By dynamically scaling the semantic velocity difference $\Delta V(t_i)$, our method achieves coarse-to-fine editing: aggressive modification during active semantic integration and gentle refinement (or attenuation) once the target concept is realized. Crucially, this adaptive gain operates independently of any pre-specified editing window—it is driven solely by semantic feedback.

\subsection{Verifying: Dynamic Windowing and Dual-Level Feedback}
\label{subsec:verifying}

The \textbf{Verifying} module enables closed-loop control by determining \textit{when} and \textit{how long} to edit, effectively realizing a \textbf{dynamic editing window}. Instead of committing to a fixed range of timesteps, we evaluate the editing progress at regular intervals—every $k=5$ steps—and automatically decide whether to continue or terminate the current trajectory.

At each verification point $t$, we decode the current latent $Z_t^{\text{UE}}$ into an image $I_t = \mathcal{G}_{\text{dec}}(Z_t^{\text{UE}})$ and prompt the frozen VLM to produce two signals:
\begin{itemize}
    \item A global alignment score $s_t$ (via CLIP-Sim with $C_{\text{tar}}$);
    \item A task-completion score $p_t$ (via language prompting, e.g., ``Rate alignment with the instruction on a 0–1 scale'').
\end{itemize}
    
While these signals are consumed by the Editing module to adjust gain (Sec.~\ref{subsec:editing}), their another role is to \textbf{gate the editing window}. We implement \textbf{early stopping}: if both $s_t > 0.85$ and $p_t > 0.9$ for two consecutive feedback points, we halt denoising immediately. This yields a data-adaptive editing window $[T, t_{\text{stop}}]$, which may end far before $t=0$.

Beyond intra-run control, Verifying also supports \textbf{inter-cycle refinement}. If the final output fails to meet fidelity thresholds ($s_0 < 0.85$ or $p_0 < 0.9$), we prompt the VLM to generate a discrepancy analysis (e.g., ``the sunset is too dim and lacks clouds''), which is converted into a corrective instruction $q_{\text{new}}$. The entire UEV loop then restarts from the best intermediate latent $Z_{t^*}^{\text{UE}}$, where $t^* = \arg\max_t s_t$.

This dual-level feedback—\textit{early stopping within a trajectory} and \textit{iterative retry across trajectories}—ensures that UniEdit-I  edits \textit{smartly} and \textit{completely}, transforming the VLM into an active conductor of the editing process.

\begin{figure*}[t]
  \centering
  \includegraphics[width=\textwidth]{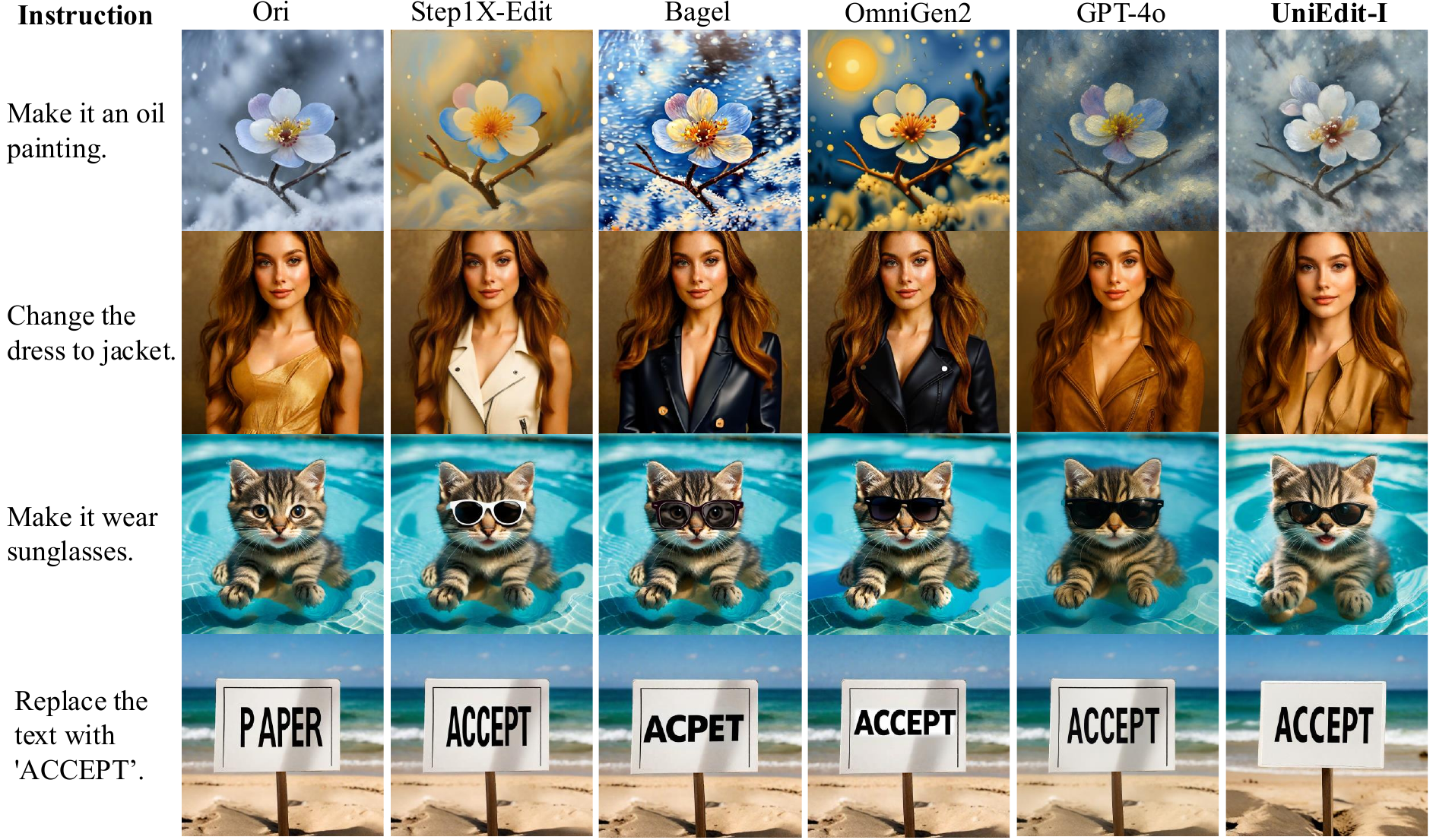}
  \caption{Qualitative comparisons. We compare UniEdit-I with recent unified VLMs across various editing tasks.}
  \label{fig:comparison}
\end{figure*}

\section{Experiments}
\label{sec:experiments}

We conduct extensive experiments to evaluate UniEdit-I across multiple dimensions: (1) its ability to generate artifact-free trajectories in semantic space, (2) its performance on the large-scale GEdit-Bench benchmark, (3) its adaptability to diverse editing tasks via feedback tuning, (4) the necessity of the unified VLM's semantic space, and (5) the efficacy and efficiency of the UEV loop. All experiments are conducted using the publicly available \textbf{BLIP3-o-8B} as the unified VLM, with a diffusion time step $T = 30$, classifier-free guidance (CFG) scales of $2.0$ (source) and $5.5$ (target), and a maximum of 3 UEV iterations. Unless otherwise specified, all metrics are reported on the English subset of GEdit-Bench~\cite{gedit}.

\subsection{Quantitative Evaluation on GEdit-Bench}
\label{subsec:quant_eval}

\begin{table}[ht]
  \centering
  \small
  \caption{Evaluation on GEdit-Bench-EN (Full set).}
  \label{tab:gedit-bench-en}
  \begin{tabular}{llccc}
    \toprule
    \textbf{Type} & \textbf{Model} 
      & \multicolumn{3}{c}{\textbf{GEdit-Bench-EN (Full set)} $\uparrow$} \\
    \cmidrule(lr){3-5}
    & & \textbf{G\_SC} & \textbf{G\_PQ} & \textbf{G\_O} \\
    \midrule
    \multirow{1}{*}{Private}
      & GPT-4o     & \textbf{7.85} & \textbf{7.62} & \textbf{7.53} \\
    \midrule
    \multirow{6}{*}{Open-source}
      & Instruct-Pix2Pix           & 3.58 & 5.49 & 3.68 \\
      & OmniGen                    & 5.96 & 5.89 & 5.06 \\
      & Step1X-Edit                & 7.09 & 6.76 & 6.70 \\
      & BAGEL                      & \textbf{7.36} & 6.83 & 6.52 \\
      & OmniGen2                   & 7.16 & 6.77 & 6.41 \\
      \rowcolor{highlight}
      & \textbf{UniEdit-I}(ours)   & 7.16 & \textbf{7.40} & \textbf{7.06} \\
    \bottomrule
  \end{tabular}
\end{table}

We evaluate UniEdit-I on the GEdit-Bench benchmark~\cite{gedit}, which contains diverse image editing tasks with human-annotated English and Chinese instructions, covering object replacement, material modification, scene transformation, and text change. We report results using the official \textbf{VIEScore} evaluation system, which combines three metrics: Semantic Quality (SQ), Perceptual Quality (PQ), and Overall Score (O), all scored on a 10-point scale.

As shown in Table~\ref{tab:gedit-bench-en}, UniEdit-I achieves state-of-the-art performance, even surpassing open-source baselines including Step1X-Edit, BAGEL, and OmniGen2. Notably, our method achieves a competitive Overall Score of \textbf{7.06}, approaching the performance of proprietary models like GPT-4o (7.53), despite operating without any fine-tuning or additional training. Detailed per-category results, provided in Appendix~A, further show that our method performs particularly well on background change, subject replacement, and color alteration tasks, while exhibiting a noticeable drop in performance on the text change task.

\subsection{Qualitative Comparison and Task-Specific Strengths}
\label{subsec:qualitative}

\begin{figure}[ht]
    \centering
    \includegraphics[width=0.6\textwidth]{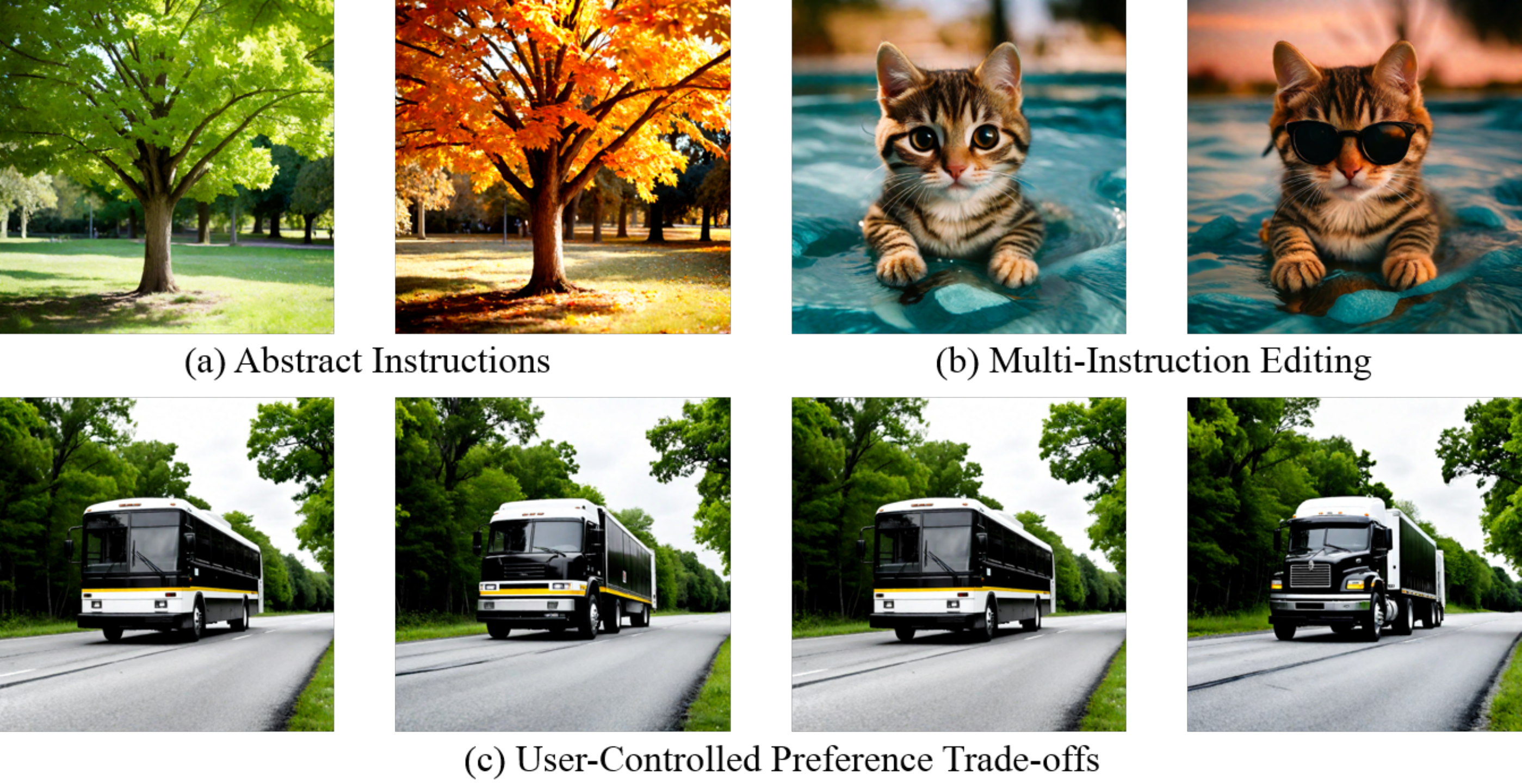}
    \caption{Task-specific editing via feedback tuning: (a) abstract instruction, (b) multi-instruction, (c) user preference trade-offs.}
    \label{fig:task_specific}
\end{figure}

In Figure~\ref{fig:task_specific}, we further highlight its adaptability through feedback prompt tuning across three challenging editing scenarios without changing the core algorithm.

\textbf{Abstract instructions.} For vague prompts like ``make it feel autumnal'', our \textit{Understanding} stage grounds the request into explicit targets (e.g., ``red/yellow leaves, fallen foliage''). As shown in Figure~\ref{fig:task_specific}(a), UniEdit-I produces a coherent seasonal transformation.

\textbf{Multi-instruction editing.} For compound edits (e.g., ``put sunglasses on it and change the time of day to dusk''), we prompt the VLM to evaluate sub-tasks independently and combine their feedback. Figure~\ref{fig:task_specific}(b) shows that UniEdit-I satisfies both conditions.

\textbf{User-controlled trade-offs.} By modifying the verification prompt (e.g., ``prioritize semantic  accuracy'' vs. ``preserve original features''), users can explicitly control the balance between semantic fidelity and structural consistency. Figure~\ref{fig:task_specific}(c) illustrates two valid outcomes from the same input, showcasing controllable, intent-aware editing.

\subsection{Why Semantic Space Matters: Artifact-Free Trajectories Enable Stable Feedback}
\label{subsec:artifact_and_stability}
To validate our core hypothesis—that a unified VLM's semantic latent space is essential for reliable closed-loop editing—we compare intermediate states generated by FlowEdit in VAE (FLUX~\cite{flux}), and CLIP (BLIP3-o) spaces. We evaluate both \textit{visual quality} and \textit{feedback stability}, as both are critical for effective verification.

First, we assess \textbf{visual artifacts} using a frozen VLM with a custom prompt: \textit{``Rate the presence of ghosting, object deformation, or unnatural textures on a scale of 1–10, where 1 means severe artifacts and 10 means visually clean.''} As shown in Table~\ref{tab:artifact_and_stability}, CLIP space produces consistently clean intermediates (score: 8.10), while VAE spaces suffer from severe distortions (5.35).

Second, we measure the \textbf{temporal stability} of semantic feedback by computing the standard deviation of $\text{CLIP-Sim}(I_t, C_{\text{tar}})$ across timesteps. In pixel/VAE space, artifacts cause erratic fluctuations in alignment scores (std = 0.025), making downstream gain control unreliable. In contrast, CLIP space yields smooth, stable feedback (std = 0.063), enabling precise modulation of editing intensity.

\begin{table}[ht]
  \centering
  \caption{Visual artifact severity and feedback stability across latent spaces (mean $\pm$ std over 100 samples).}
  \label{tab:artifact_and_stability}
  \begin{tabular}{lcc}
    \toprule
    \textbf{Latent Space} & \textbf{Artifact Score} ($\uparrow$) & \textbf{Feedback stability} ($\downarrow$) \\
    \midrule
    VAE (FLUX) & 5.35 $\pm$ 1.02 & 0.063 \\
    CLIP (BLIP3-o) & \textbf{8.10 $\pm$ 0.53} & \textbf{0.025} \\
    \bottomrule
  \end{tabular}
\end{table}

Figure~\ref{fig:semantic} visualizes this contrast: VAE space exhibits large oscillations in alignment scores due to transient artifacts, while CLIP space maintains a smooth, monotonic progression toward the target. More examples can be found in Appendix.

\subsection{Effectiveness of the UEV Loop}
\label{subsec:uev_effectiveness}

\subsubsection{Ablation on Dynamic Gain Mechanism}
\label{subsubsec:gain_ablation}
We ablate the design of the gain schedule in CLIP space under the same UEV framework (with dynamic windowing and 3-refinement limit). As shown in Table~\ref{tab:gain_ablation}, using a fixed gain ($\alpha_t = 1.0$)—as in FlowEdit—leads to over-editing and high output variance. Linear decay ($\alpha_t = 1.0 - 0.03t$) mitigates drift slightly but still lacks responsiveness to semantic progress. Removing the task-completion score $p_t$ (i.e., using only $\Delta s_t$) results in failing to reach the optimal performance (highest scores) and occasional over-modification, as the system cannot reliably detect when the edit goal is fully satisfied. In contrast, our full dynamic gain achieves the highest scores across all metrics and the lowest standard deviation, confirming that both alignment dynamics ($\Delta s_t$) and completion awareness ($p_t$) are essential for stable, high-fidelity editing.

\begin{table}[ht]
\centering
\small
\caption{Ablation on gain scheduling strategies (GEdit-Bench-EN, full set). All methods use CLIP space and dynamic windowing; only gain logic varies.}
\label{tab:gain_ablation}
\begin{tabular}{lccc}
\toprule
\textbf{Gain Strategy} & \textbf{SQ} $\uparrow$ & \textbf{PQ} $\uparrow$ & \textbf{O} $\uparrow$ \\
\midrule
Fixed gain ($\alpha_t = 1.0$) & 5.87  & 7.39 & 5.66 \\
Linear decay & 6.16  & 7.42 & 5.97 \\
Dynamic gain (no $p_t$) & 6.73  & 7.38 & 6.77 \\
\textbf{Full dynamic gain (ours)} & \textbf{7.16} & \textbf{7.40} & \textbf{7.06} \\
\bottomrule
\end{tabular}
\end{table}

\subsubsection{Feedback Convergence Behavior.}
\label{subsubsec:convergence}
With feedback every 5 steps and up to 3 iterations, UniEdit-I converges efficiently. As shown in Table~\ref{tab:convergence_steps}, most tasks reach their optimal output well before the full denoising trajectory, thanks to the effective early-stopping mechanism guided by real-time semantic verification.
\begin{table}[ht]
\centering
\small
\caption{Convergence distribution over GEdit-Bench-EN}
\label{tab:convergence_steps}
\begin{tabular}{lcc}
\toprule
\multicolumn{2}{c}{\textbf{Convergence Stage}}  & \textbf{\% of Samples} \\
\midrule
\multirow{4}{*}{First iteration} 
    & $t=30$                       & 10.2\% \\
    & $t=25$                       & 18.0\% \\
    & $t=20$                       & 26.6\% \\
    & $t=15$                       & 23.7\% \\
    & $t=10$ or earlier            & 19.0\% \\
    \cmidrule(lr){2-3}
    & \textbf{Total in first pass}              & \textbf{97.6\%} \\
\midrule
\multicolumn{2}{c}{After 1 refinement iteration}  & 2.5\% \\
\midrule
\multicolumn{2}{c}{\textbf{Overall convergence}} & \textbf{100\%} \\
\bottomrule
\end{tabular}
\end{table}

\section{Limitations and Conclusion}
In conclusion, UniEdit-I establishes a new paradigm for image editing by treating the unified VLMs not as a static evaluator, but as an active, in-process conductor of semantic refinement—through a training-free, closed-loop mechanism that dynamically steers editing trajectories in the CLIP semantic space. By embedding real-time feedback into the editing process and leveraging the intrinsic alignment of unified VLMs between text and image representations, our method achieves state-of-the-art semantic fidelity without fine-tuning, demonstrating that robust, intention-aligned editing emerges not from scale or data, but from structured reflection within the model’s own latent space.

\textbf{Limitations: }Our method still has several limitations. our method inherits the semantic representation space of the underlying unified VLM (e.g., BLIP3-o); thus, any biases or coverage gaps in the pretrained vision–language model (e.g., rare concepts or fine-grained attributes) may propagate into the editing process. This may explain our low scores on the text change task.

\clearpage

\bibliographystyle{plainnat}
\bibliography{main}

\clearpage


\newpage
\appendix
\section*{Appendix}

\section{Additional Settings and Results}
\subsection{CLIP Score Setting}
we use CLIP-ViT-B/32 for the experiments on CLIP score and CLIP feature cosine similarity.

\subsection{GEdit-Bench Categorize Results}
\paragraph{GEdit-Bench.}
We adopt \textbf{GEdit-Bench}, a highly authentic and representative benchmark, as the core framework for systematically evaluating our proposed image editing model. Developed by the Step1X-Edit research team, GEdit-Bench has emerged as a widely recognized and authoritative benchmark in the field of text-guided image editing. It comprises 606 meticulously curated test samples, constructed through a multi-stage pipeline. Over 1000 real-world image editing requests were initially collected from platforms such as Reddit and subsequently filtered to remove redundancy and ensure relevance. Each sample includes an original image, a natural language editing instruction, and a detailed description of the desired outcome, providing rich contextual information for comprehensive evaluation. To enable fine-grained analysis, all samples were manually annotated and categorized into 11 distinct editing task types—such as object manipulation, attribute modification, and style transfer—ensuring balanced task coverage and facilitating rigorous assessment of model performance across diverse editing objectives.

\begin{table}[ht]
  \centering
  \small
  \caption{GEdit Categorize Results}
  \label{tab:classification-results}
  \begin{tabular}{lccc}
    \toprule
    \textbf{Task Type} 
    & \multicolumn{3}{c}{\textbf{GEdit-Bench-EN (Full set)} $\uparrow$} \\
    \cmidrule(lr){2-4}
                           & \textbf{G\_SC} & \textbf{G\_PQ} & \textbf{G\_O} \\
    \midrule
    \rowcolor{highlight}
    background\_change     & 8.050 & 7.650 & 7.747 \\
    color\_alter           & 7.750 & 7.675 & 7.462 \\
    material\_alter        & 6.275 & 7.300 & 6.472 \\
    motion\_change         & 7.600 & 7.550 & 7.444 \\
    ps\_human              & 7.271 & 7.543 & 7.290 \\
    style\_change          & 7.483 & 7.467 & 7.386 \\
    \rowcolor{highlight}
    subject-add            & 7.900 & 7.750 & 7.668 \\
    subject-remove         & 6.807 & 7.333 & 6.739 \\
    subject-replace        & 7.683 & 7.267 & 7.335 \\
    \rowcolor{badcase}     
    text\_change           & 4.000 & 6.434 & 4.495 \\
    \rowcolor{highlight}
    tone\_transfer         & 7.900 & 7.425 & 7.570 \\
    \midrule
    \textbf{Average}       & 7.156 & 7.399 & 7.055 \\
    \bottomrule
  \end{tabular}
\end{table}

\paragraph{GEdit-Bench Categorize Results.}
Table~\ref{tab:classification-results} reports the subjective evaluation results of GEdit on GEdit-Bench-EN (Full set) across different editing task categories. The metrics include G\_SC (semantic consistency), G\_PQ (perceptual quality), and G\_O (overall performance), all scored on a 0--10 scale, where higher is better.
Overall, most task types achieve scores in the range of 6.5--7.8, indicating that our method produces stable and high-quality editing results across diverse scenarios. Tasks such as background\_change (G\_SC=8.050, G\_O=7.747), subject-add (G\_SC=7.900, G\_O=7.668), and tone\_transfer (G\_SC=7.900, G\_SC=7.570) rank the highest, suggesting strong generalization and visual coherence for background replacement, subject addition, and style transfer edits. In contrast, the text\_change task shows significantly lower scores (G\_SC=4.000, G\_O=4.495), highlighting that text-specific edits remain challenging. this can be attributed to the limitations inherited from the underlying unified vision-language model (VLM), Since our method operates within the pre-trained representation space, it cannot fully overcome these shortcomings during the editing process. As a result, when tasked with modifying or generating precise textual content, the model struggles to produce semantically accurate or visually consistent changes, leading to the observed lower performance on the text\_change task.

\section{Effectiveness of Dynamic Gain Mechanism}

\begin{figure*}[htbp]
 \centering
 \includegraphics[width=0.7\textwidth]{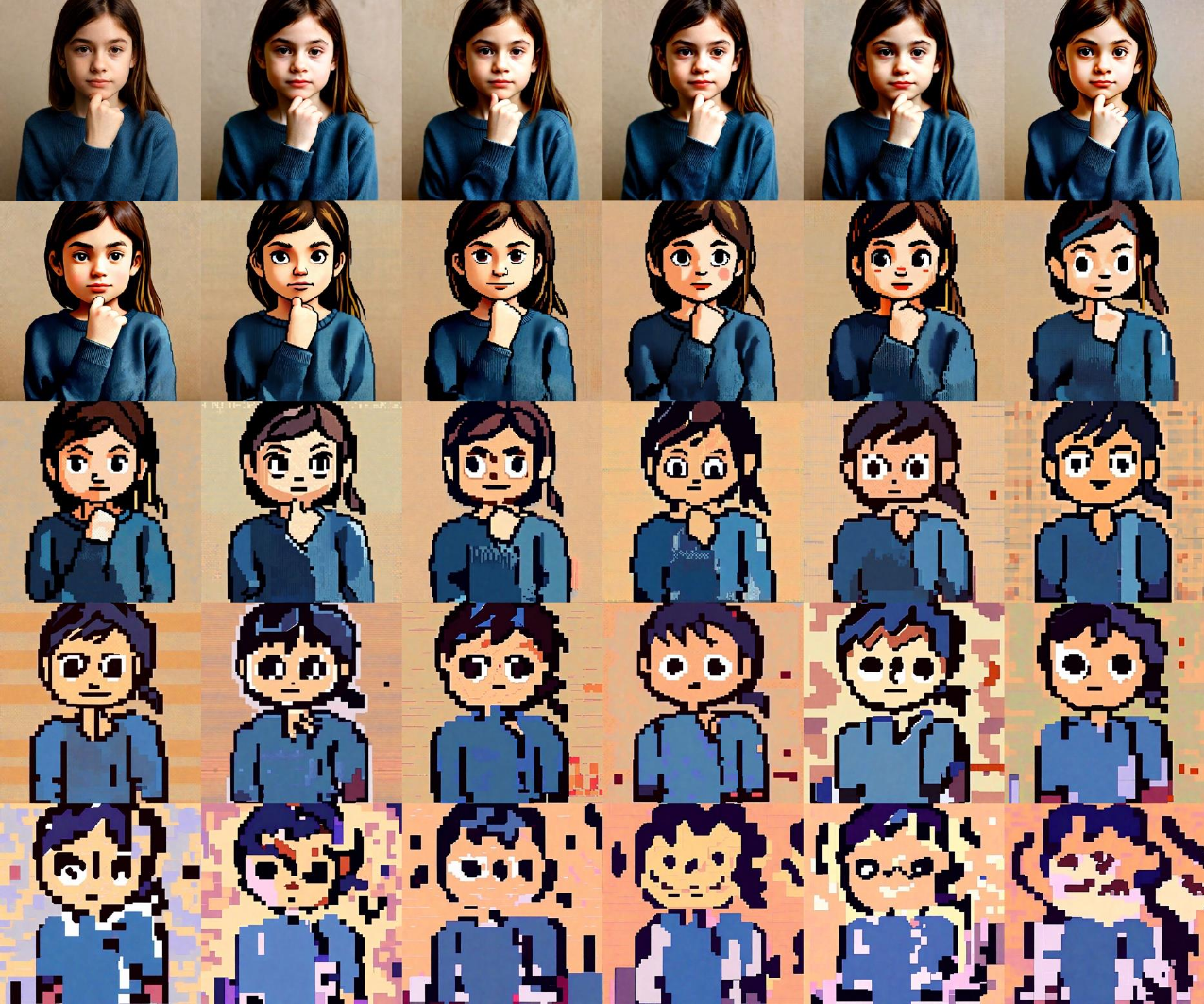}
 \caption{Example of Fixed Gain.}
 \label{fig:b1}
\end{figure*}

\begin{figure*}[htbp]
 \centering
 \includegraphics[width=0.7\textwidth]{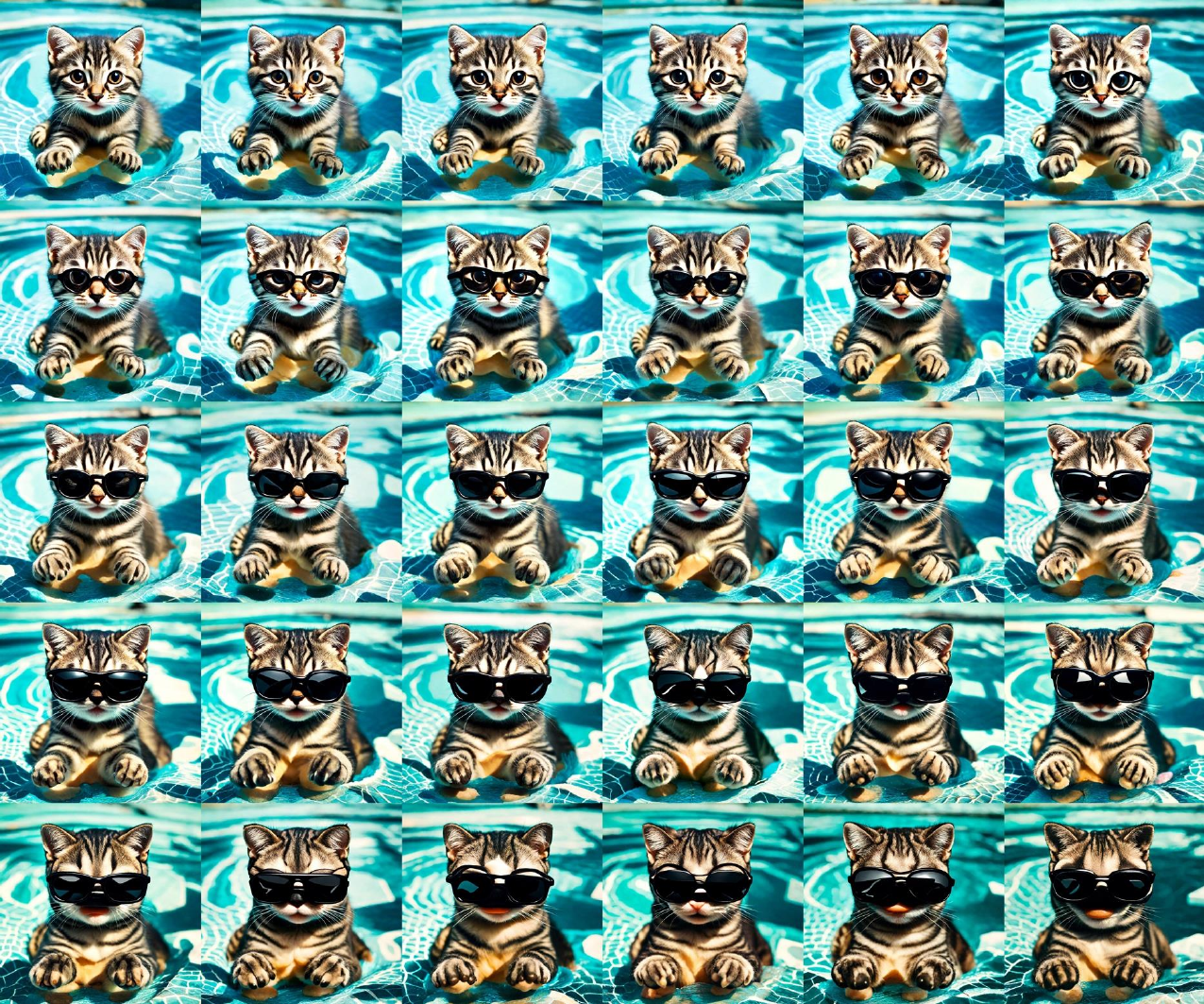}
 \caption{Example of Fixed Gain.}
 \label{fig:b3}
\end{figure*}

\begin{figure*}[htbp]
 \centering
 \includegraphics[width=0.7\textwidth]{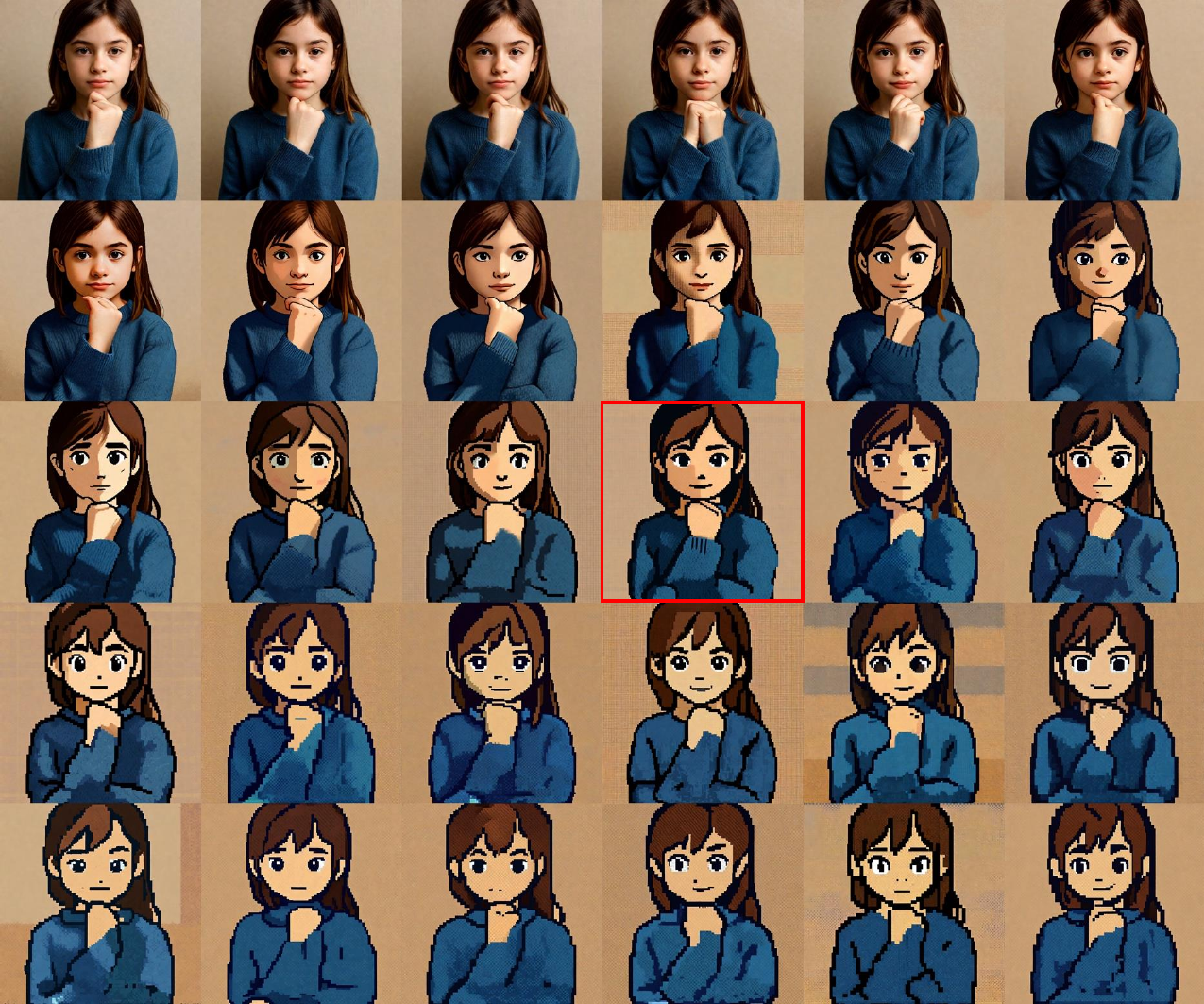}
 \caption{Example of Dynamic Gain.}
 \label{fig:b2}
\end{figure*}

\begin{figure*}[htbp]
 \centering
 \includegraphics[width=0.7\textwidth]{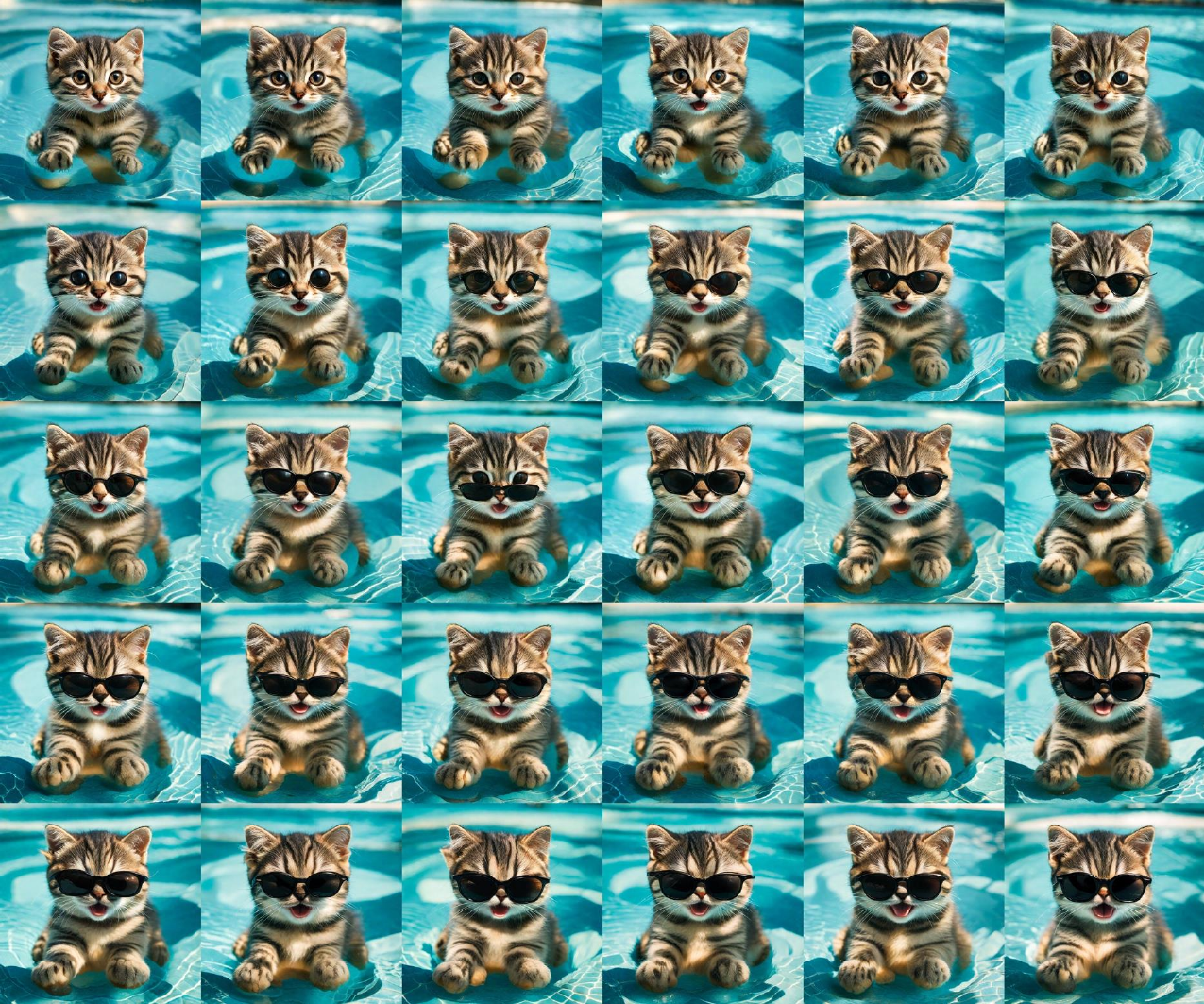}
 \caption{Example of Dynamic Gain.}
 \label{fig:b4}
\end{figure*}

Fixed gain ($\alpha_t = 1.0$) applies a constant editing strength throughout the entire process, lacking awareness of the current semantic alignment progress or whether the editing goal has been achieved. As a result, it is highly prone to \textit{over-editing} or \textit{under-editing}. Specifically, when the target has already been reached, the system continues modifying the image, introducing unnecessary perturbations to irrelevant regions and degrading structural integrity and visual consistency. Conversely, in complex tasks, fixed gain may lead to insufficient editing due to premature termination, failing to fully satisfy the user instruction. This method follows a pre-defined, static trajectory—an open-loop control mechanism—akin to ``blindly pushing forward'' without leveraging real-time feedback for dynamic correction or adaptive intensity modulation, making precise and reliable editing difficult to achieve.

In contrast, the dynamic gain mechanism fundamentally addresses this limitation by tightly coupling editing intensity with real-time semantic feedback, thereby resolving the misalignment caused by the absence of process perception. 
The gain coefficient is defined as:
\[
    \alpha_t = \alpha_{\text{base}} \cdot \sigma(\kappa_1 \Delta s_t) \cdot (1 - p_t),
\]
where $\alpha_{\text{base}} = 1.0$, $\kappa_1 = 15$, $\Delta s_t = s_t - s_{\text{prev}}$ measures the improvement in global semantic alignment since the last feedback point, $p_t \in [0,1]$ denotes the task completion score, and $\sigma(\cdot)$ is the sigmoid function. This formulation enables adaptive control: when semantic alignment is improving ($\Delta s_t > 0$), the gain is amplified to accelerate convergence; as the output approaches the target ($p_t \to 1$), the gain is gradually attenuated to prevent over-modification. Consequently, the dynamic gain significantly enhances both the stability of the editing process and the fidelity of the final output.

More importantly, this mechanism works synergistically with UniEdit-I's verification module to form a closed-loop control system. By incorporating multimodal semantic feedback from the VLM every $k=5$ diffusion steps, and combining it with an early-stopping criterion—halting the process immediately when $s_t > 0.85$ and $p_t > 0.9$ are simultaneously satisfied for two consecutive evaluations—the framework achieves intelligent ``stop-upon-success'' regulation. This enables automatic adaptation to varying task complexity: simple edits converge rapidly, while complex ones undergo progressive refinement. Crucially, no manual parameter tuning is required. As a result, UniEdit-I realizes efficient, robust, and user-friendly training-free image editing, transforming the process from rigid, open-loop execution into a responsive, self-correcting loop guided by semantic intelligence.

\section{More Examples}

\subsection{From Pixels to Semantics: Editing at the Conceptual Level}

\begin{figure*}[htbp]
 \centering
 \includegraphics[width=\textwidth]{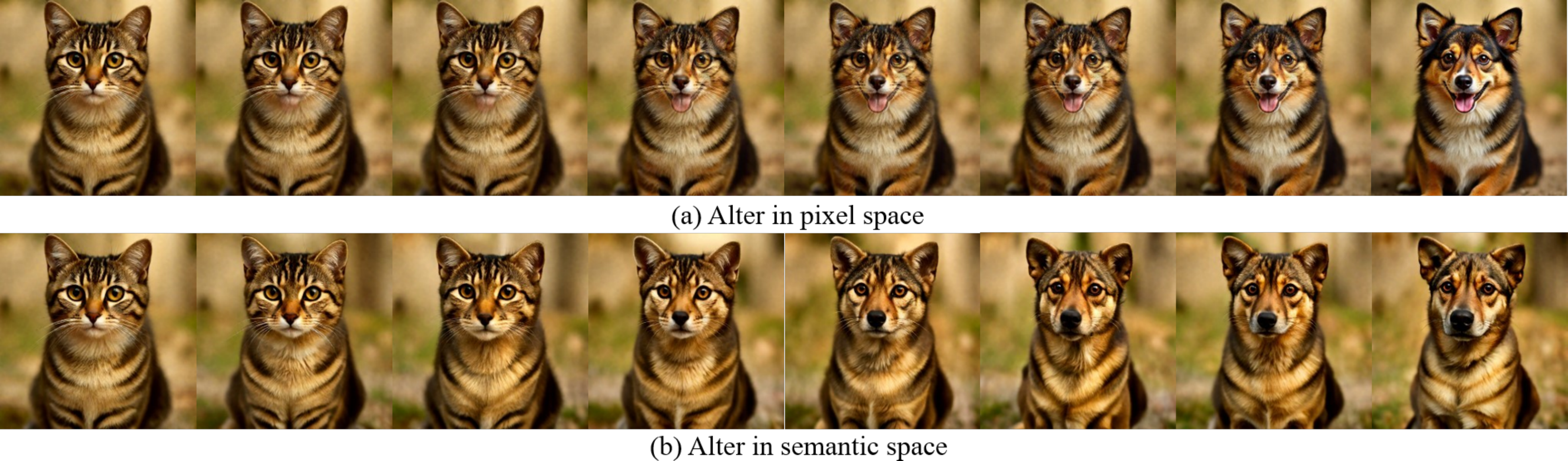}
 \caption{Comparison of (a) Editing in pixel space and (b) Editing in semantic space.}
 \label{fig:s1}
\end{figure*}

\begin{figure*}[htbp]
 \centering
 \includegraphics[width=\textwidth]{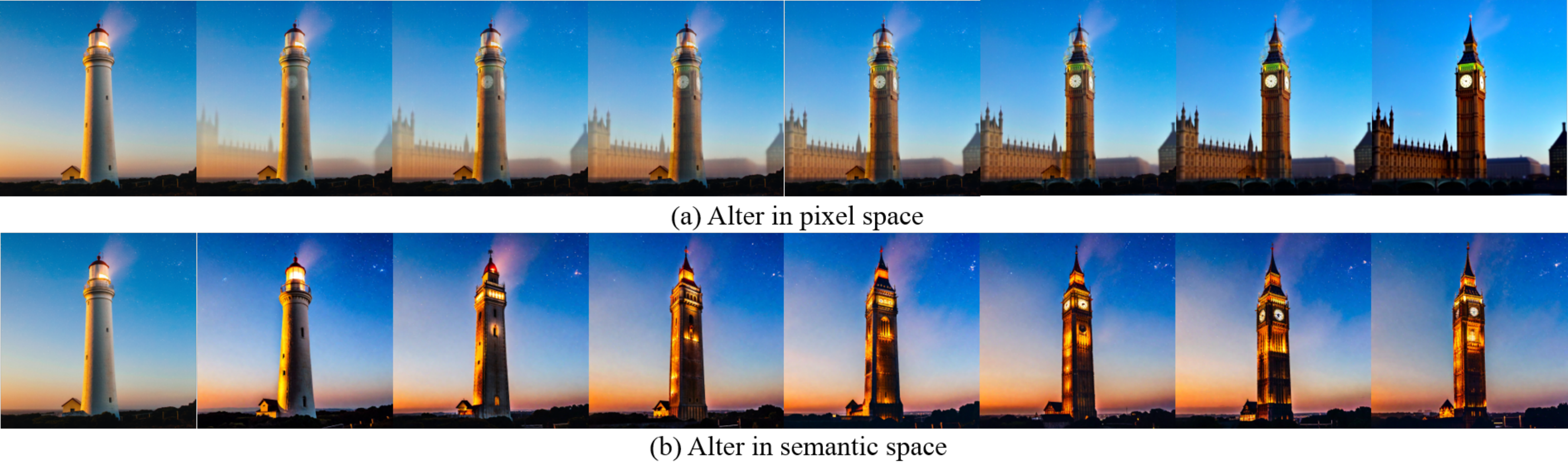}
 \caption{Comparison of (a) Editing in pixel space and (b) Editing in semantic space.}
 \label{fig:s2}
\end{figure*}

\begin{figure*}[htbp]
 \centering
 \includegraphics[width=\textwidth]{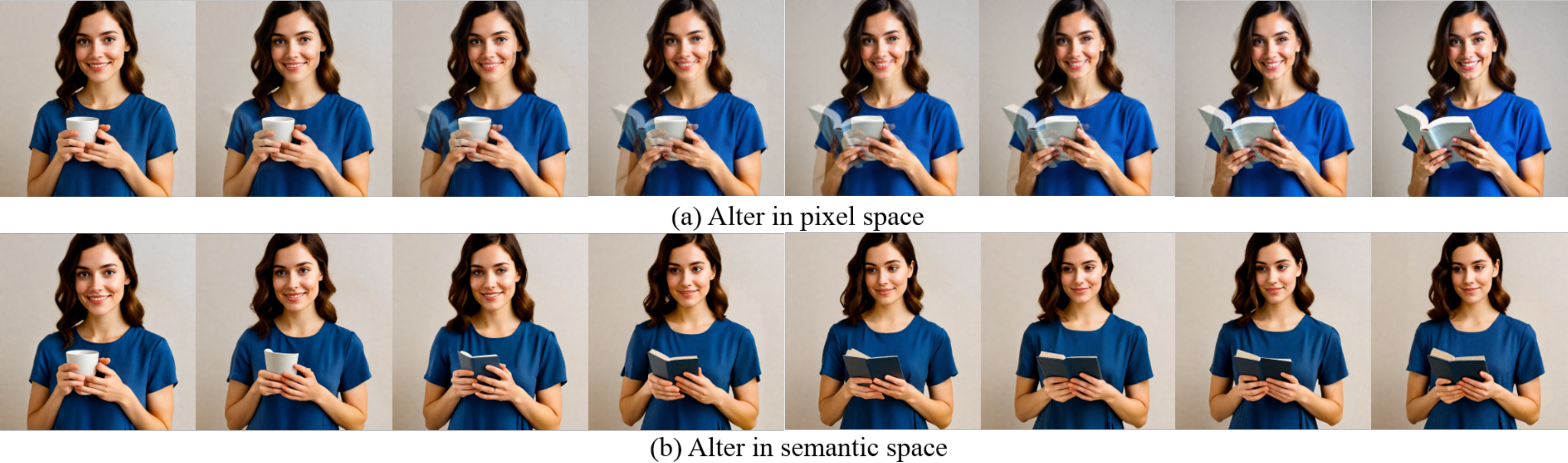}
 \caption{Comparison of (a) Editing in pixel space and (b) Editing in semantic space.}
 \label{fig:s3}
\end{figure*}

As illustrated in Figure~\ref{fig:s1}, \ref{fig:s2}, and \ref{fig:s3}, we conduct image editing at the semantic level rather than operating directly in pixel space. This paradigm shift enables a more structured, interpretable, and semantically consistent transformation process. Unlike traditional pixel-level editing—where modifications are applied by directly altering pixel intensities (Figure~\ref{fig:s1}(a), \ref{fig:s2}(a), and \ref{fig:s3}(a))—semantic editing operates on high-level representations that encode meaningful attributes such as object identity, pose, texture, and spatial relationships. By manipulating these abstract semantic features, the model adjusts the conceptual definition of the image content, ensuring that each intermediate and final output remains both visually plausible and semantically coherent (Figure~\ref{fig:s1}(b), \ref{fig:s2}(b), and \ref{fig:s3}(b)).

One of the key advantages of semantic editing lies in its the visibility and interpretability of intermediate editing steps while performing complex transformations. Pixel-level methods often suffer from artifacts such as blurring, misalignment, or structural distortions when handling edits, especially when multiple changes are required simultaneously. These approaches typically lack an understanding of scene semantics, treating images as mere grids of pixels without awareness of higher-order concepts like object hierarchy or contextual dependencies. In contrast, because semantic editing is grounded in a conceptual understanding of the image content, it can naturally interpret logical combinations of editing commands.

Furthermore, semantic editing supports progressive refinement, enabling smooth transitions from coarse structural adjustments to fine-grained details. This hierarchical editability mirrors human perception and cognitive processing, making the editing process not only more effective but also more intuitive and user-friendly. It also facilitates interpretability, as each editing step corresponds to a meaningful change in semantic space, allowing users to understand and trace the transformation path.

While pixel-level methods remain useful for local retouching or noise-level adjustments, semantic editing offering greater robustness, consistency, and alignment with user intent.

\subsection{More Editing Eesults}

In conclusion, UniEdit-I establishes a new paradigm for image editing by treating unified vision-language models (VLMs) not as static evaluators, but as active, in-process conductors of semantic refinement. Through a training-free, closed-loop mechanism, our method dynamically steers editing trajectories in the CLIP semantic space, using real-time feedback to iteratively align visual outputs with textual instructions. By exploiting the intrinsic cross-modal alignment of VLMs, UniEdit-I achieves state-of-the-art semantic fidelity without any fine-tuning, demonstrating that robust, intention-aligned editing stems not from model scale or massive curated datasets, but from structured, self-guided reasoning within the latent space. This framework naturally supports a broad spectrum of edits—from coarse structural changes to fine-grained attribute control—and enables complex, multi-instruction compositions within a single coherent process. Unlike black-box end-to-end generators, our approach offers transparent, step-by-step interpretability, clearly revealing how semantic concepts are progressively realized across the editing loop.

We present additional visual results in the supplementary materials, further showcasing UniEdit-I’s versatility across diverse tasks. These examples highlight its consistent ability to produce high-fidelity, artifact-free outputs while faithfully adhering to nuanced or compound language prompts, all without task-specific adaptation or retraining. The results reinforce that semantic editing via closed-loop VLM guidance is both powerful and generalizable.

\begin{figure*}[htbp]
 \centering
 \includegraphics[width=0.75\textwidth]{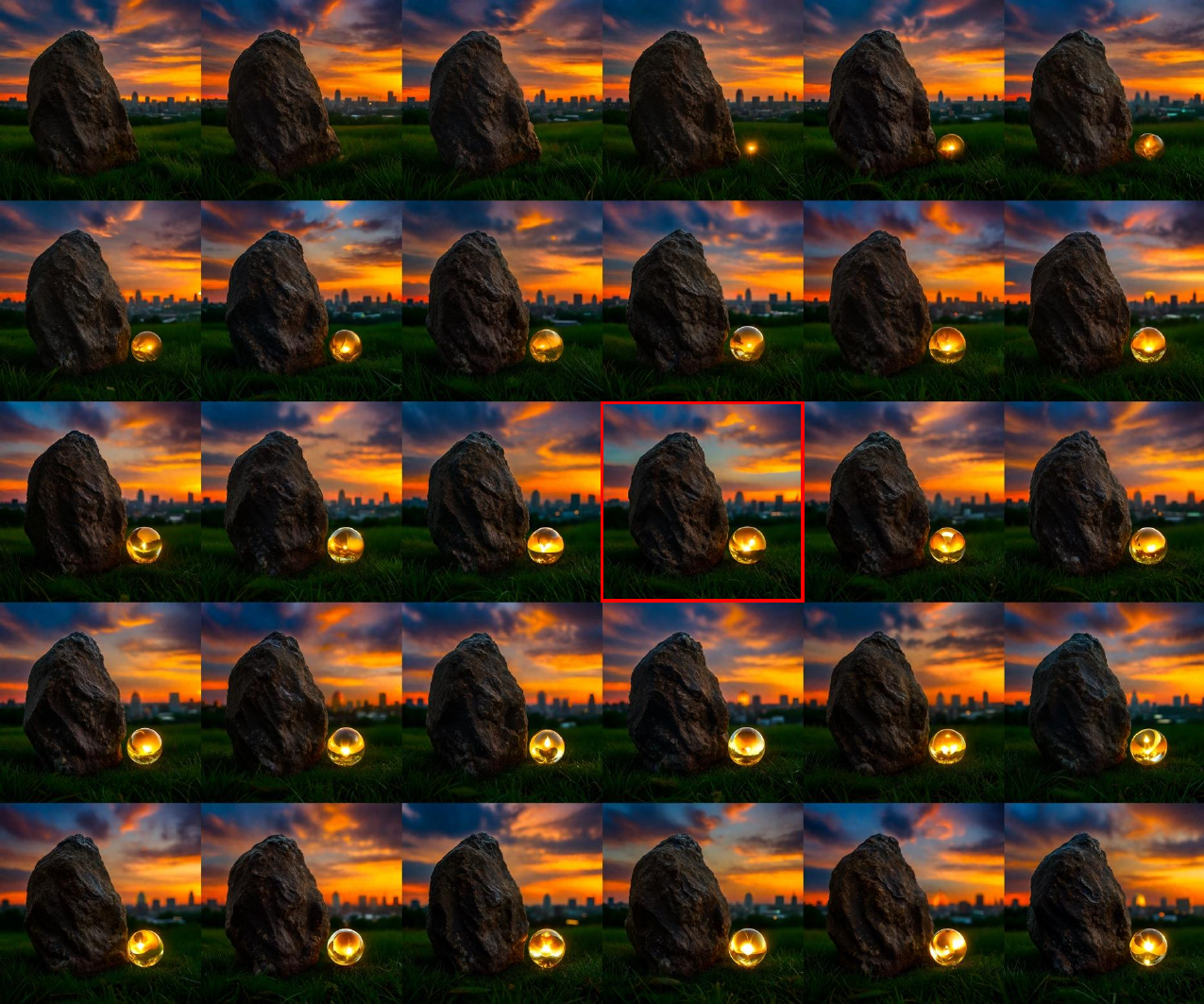}
 \caption{Subject Add Task.}
 \label{fig:c7}
\end{figure*}

\begin{figure*}[htbp]
 \centering
 \includegraphics[width=0.75\textwidth]{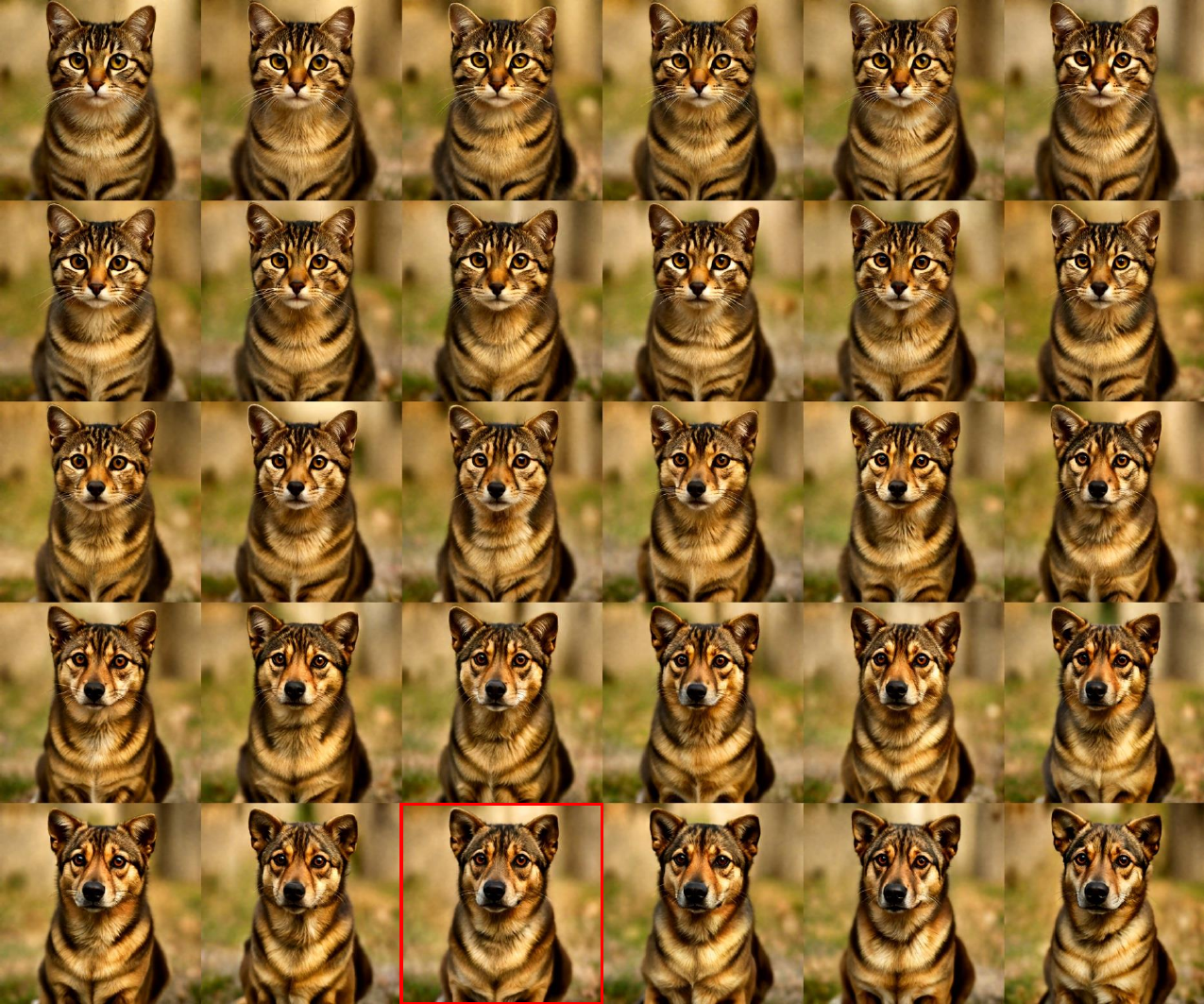}
 \caption{Subject Change Task.}
 \label{fig:c2}
\end{figure*}

\begin{figure*}[htbp]
 \centering
 \includegraphics[width=0.75\textwidth]{fig/c9.pdf}
 \caption{Color Alter Task.}
 \label{fig:c4}
\end{figure*}

\begin{figure*}[htbp]
 \centering
 \includegraphics[width=0.75\textwidth]{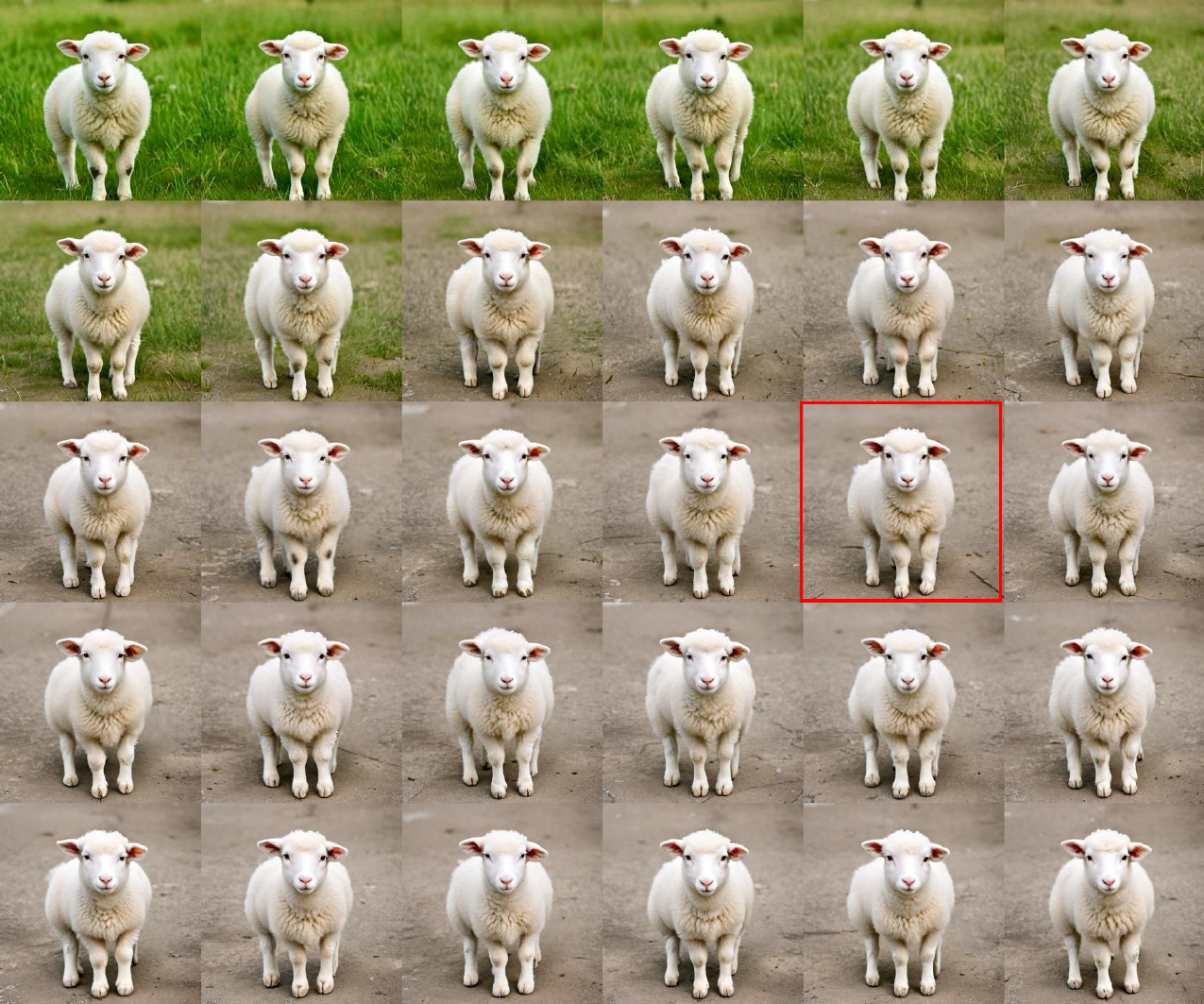}
 \caption{Background Change Task.}
 \label{fig:c1}
\end{figure*}

\begin{figure*}[htbp]
 \centering
 \includegraphics[width=0.75\textwidth]{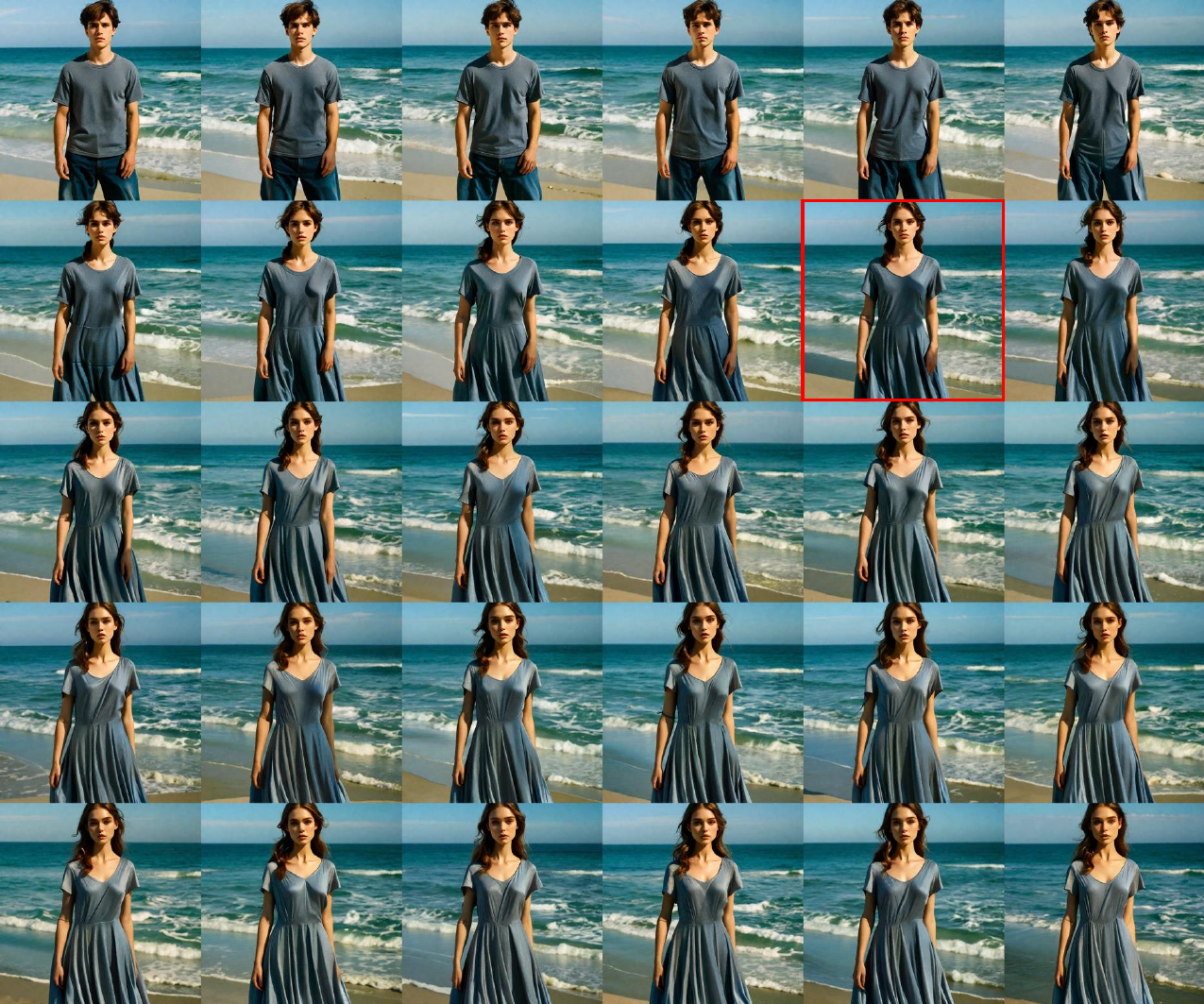}
 \caption{Attribute Change Task.}
 \label{fig:c5} 
\end{figure*}

\begin{figure*}[htbp]
 \centering
 \includegraphics[width=0.75\textwidth]{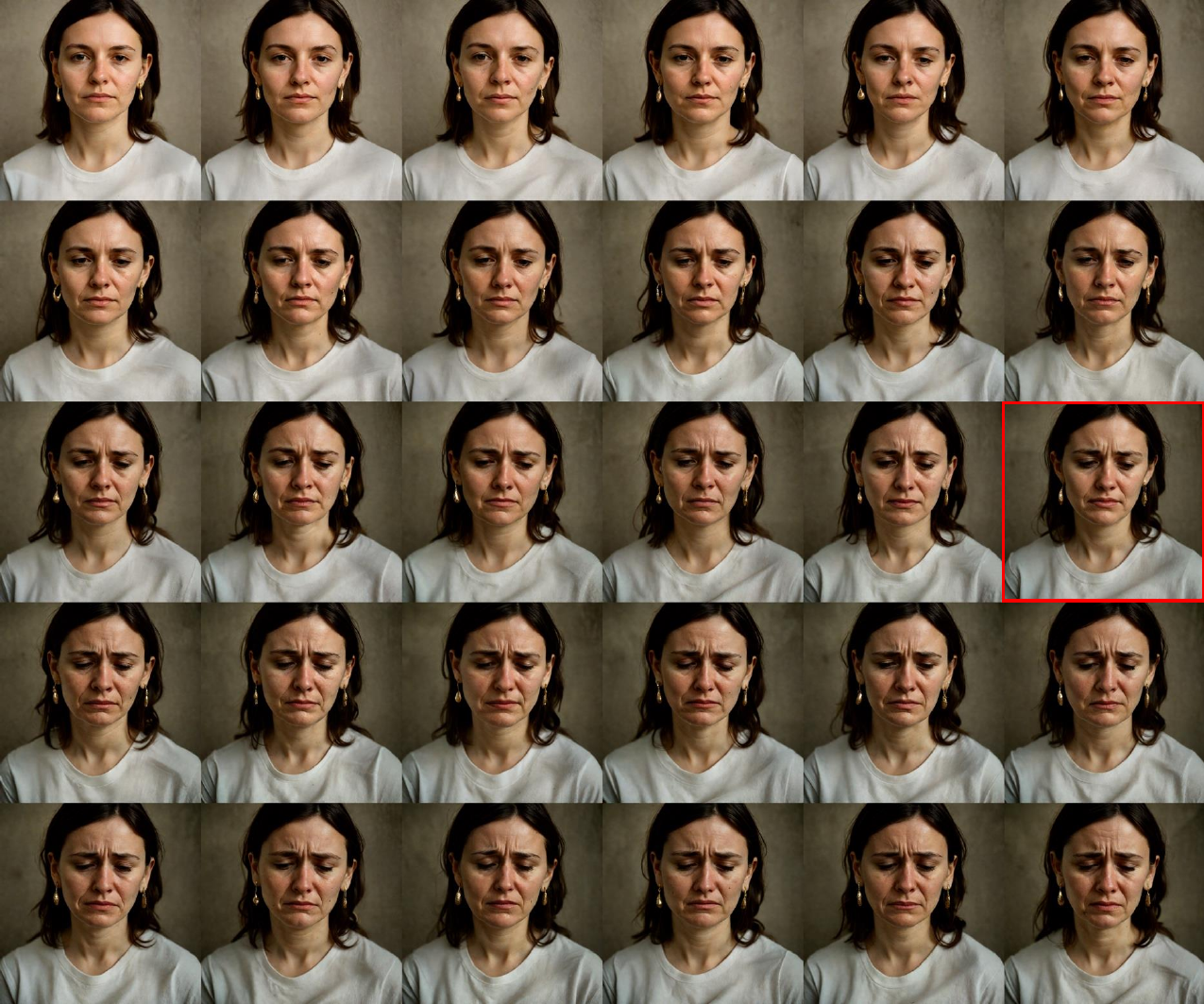}
 \caption{Motion Change Task.}
 \label{fig:c6}
\end{figure*}

\begin{figure*}[htbp]
 \centering
 \includegraphics[width=0.75\textwidth]{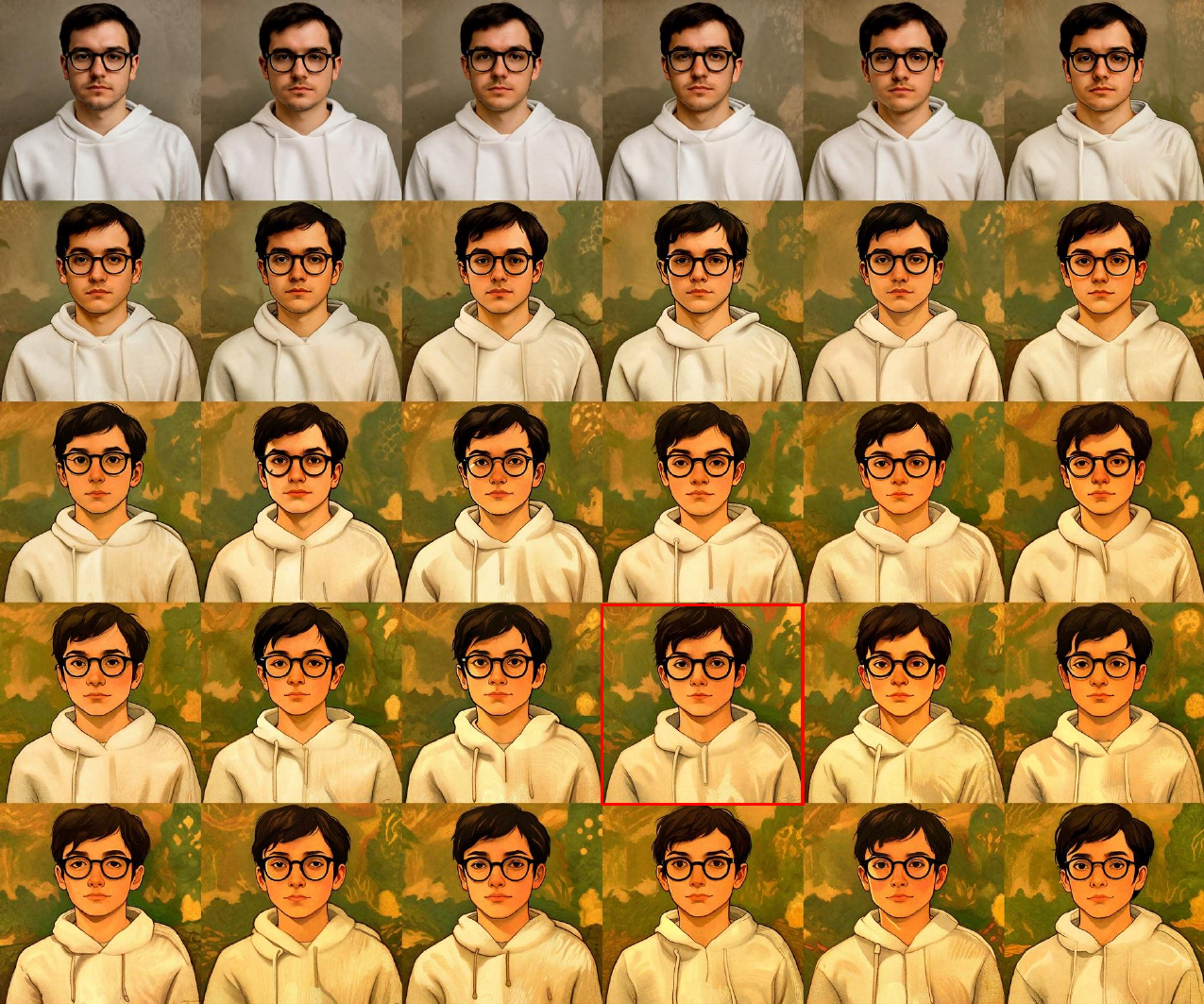}
 \caption{Style Change Task.}
 \label{fig:c3}
\end{figure*}

\begin{figure*}[htbp]
 \centering
 \includegraphics[width=0.75\textwidth]{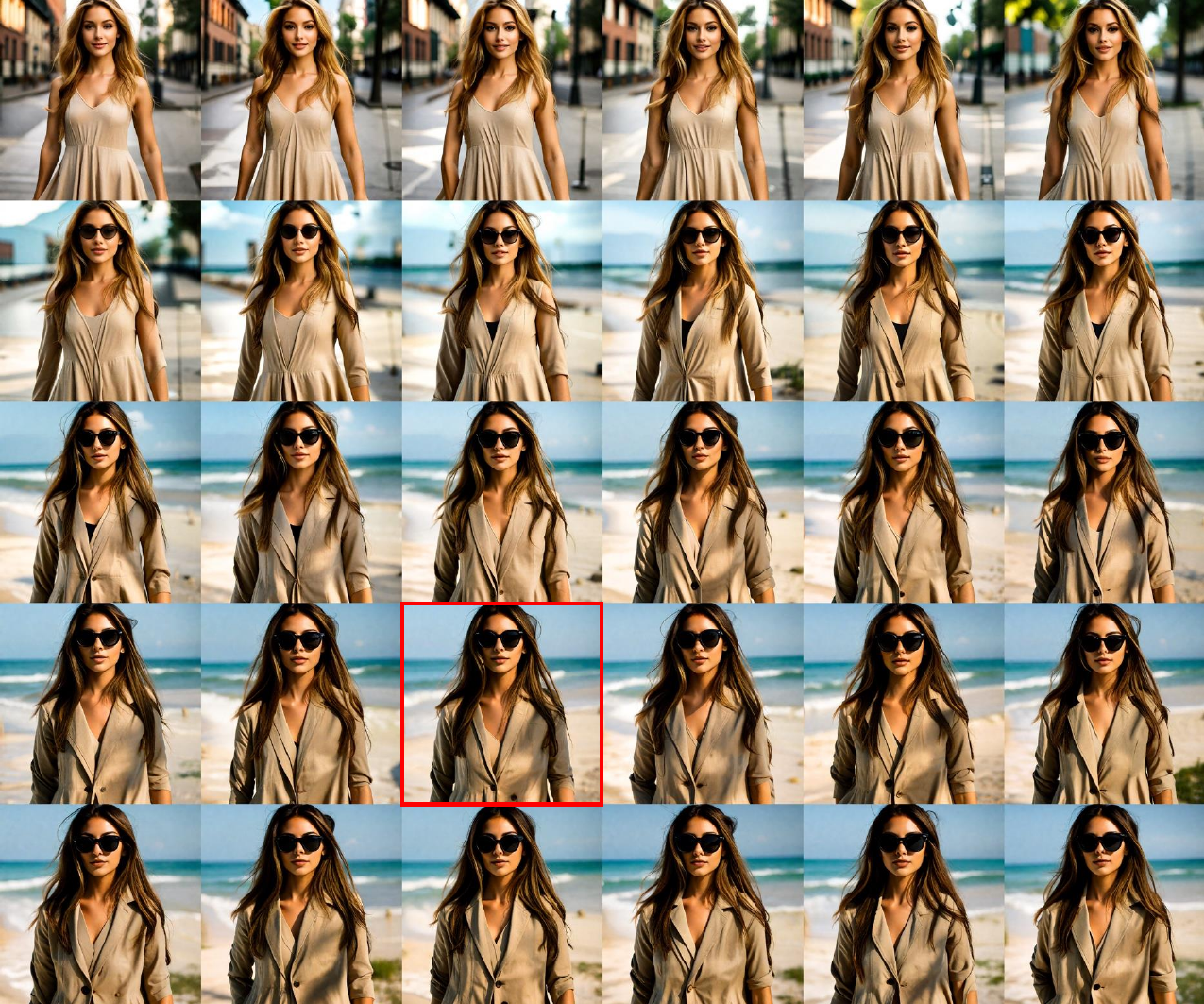}
 \caption{Multi-Task Editing.}
 \label{fig:c8}
\end{figure*}

\end{document}